\documentclass{article}
\usepackage[T1]{fontenc}
\usepackage{microtype}
\usepackage{graphicx}
\usepackage{booktabs}
\usepackage{longtable}
\usepackage{tabularx}
\usepackage{array}
\usepackage{amsmath}
\usepackage{needspace}
\usepackage{hyperref}

\usepackage[accepted]{mlsys2025}
\hypersetup{
  hypertexnames=false,
  pdfsubject={MLSys 2026 FlashInfer AI Kernel Generation Contest}
}
\makeatletter
\renewcommand{\Notice@String}{}
\renewcommand{\mlsyscorrespondingauthor}[2]{%
\ifdefined\isaccepted
  \ifdefined\mlsyscorrespondingauthor@text
    \g@addto@macro\mlsyscorrespondingauthor@text{;\\#1 \textless{}#2\textgreater{}}%
  \else
    \gdef\mlsyscorrespondingauthor@text{#1 \textless{}#2\textgreater{}}%
  \fi
\else
  \gdef\mlsyscorrespondingauthor@text{Anonymous Author \textless{}anon.email@domain.com\textgreater{}}%
\fi
}
\renewcommand{\printAffiliationsAndNotice}[1]{%
{\let\thefootnote\relax\footnotetext{\hspace*{-\footnotesep}\raggedright%
\ifdefined\mlsyscorrespondingauthor@text%
Correspondence to: \mlsyscorrespondingauthor@text.%
\else
{\bf AUTHORERR: Missing \textbackslash{}mlsyscorrespondingauthor.}%
\fi
}}}
\makeatother

\mlsystitlerunning{Harness Engineering for LLM-Driven GPU Kernel Generation}

\newcolumntype{Y}{>{\raggedright\arraybackslash}X}
\newcommand{\code}[1]{\texttt{#1}}


\makeatletter
\newcommand{\restoreMlsysRunningHead}{\chead{\small\bfseries\@mlsystitlerunning}}
\makeatother
\begin{document}
\frenchspacing
\twocolumn[
\mlsystitle{Harness Engineering for LLM-Driven GPU Kernel Generation}

\mlsyssetsymbol{baidu}{}

\begin{mlsysauthorlist}
\mlsysauthor{Yue Shui}{baidu}
\mlsysauthor{Chenyu Ma}{baidu}
\mlsysauthor{Hangfei Xu}{baidu}
\mlsysauthor{Shengzhao Wen}{baidu}
\mlsysauthor{Yanpeng Wang}{baidu}
\end{mlsysauthorlist}

\mlsyscorrespondingauthor{Yue Shui}{shuiyue@baidu.com}
\mlsyskeywords{Machine Learning Systems, GPU Kernel Optimization, LLM Agents, FlashInfer}

{\centering
\small Baidu, Inc.\par
\small\textbf{GitHub: }\href{https://github.com/syhya/mlsys26-flashinfer-contest}{github.com/syhya/mlsys26-flashinfer-contest}\par}
\vskip 0.18in

\begin{abstract}
Large language models (LLMs) can assist GPU kernel generation, but their practical effectiveness depends on whether generated code can be reliably constrained, validated, profiled, and selected. This paper presents a harness-centered system for LLM-driven GPU kernel optimization in the MLSys 2026 FlashInfer AI Kernel Generation Contest on NVIDIA Blackwell B200 GPUs. The system separates an evaluation harness from a profile-backed optimization controller: the harness enforces compilation, correctness, official-aligned timing, and artifact archival, while the controller turns profiler and workload evidence into bounded candidate-generation decisions. Human-authored skills capture operator constraints, references, profiling procedures, and promotion rules, while Codex and Claude Code agents generate candidate kernels inside those constraints. Across five operator definitions, the retained official-aligned artifacts achieved mean-latency speedups over supplied FlashInfer baselines of 1.62$\times$, 18.05$\times$, 29.68$\times$, 1.12$\times$, and 13.70$\times$. The Agent-Assisted kernels outperform the Full-Agent artifacts across the evaluated definitions, indicating that expert-provided optimization directions, high-quality references, and workload context remain critical for reliable AI-driven kernel optimization.
\end{abstract}
]
\restoreMlsysRunningHead

\printAffiliationsAndNotice{}

\section{Introduction}

Efficient LLM serving is increasingly determined by specialized runtime and kernel decisions rather than by the model graph alone, as illustrated by structured generation runtimes, paged key-value (KV) cache serving systems, IO-aware attention kernels, and production GPU inference stacks~\citep{sglang,pagedattention,flashattention,tensorrtllm}. Operators such as MoE routing, sparse attention, top-k index construction, and recurrent state updates expose irregular shapes, mixed precision, short decode regimes, and long-context regimes. The contest definitions reflect mechanisms used in DeepSeekMoE, DeepSeek-V3.2 sparse attention, and Gated Delta Networks~\citep{deepseekmoe,deepseekv32,gdn}. The FlashInfer AI Kernel Generation Contest turns these pressures into a controlled systems benchmark by evaluating submitted kernels for correctness and latency on NVIDIA B200 GPUs against FlashInfer baselines~\citep{flashinfer,flashinferbench,contest}.

LLM-based coding agents can broaden the implementation search space, as KernelBench and KernelEvolve show for efficient GPU or accelerator kernel generation~\citep{kernelbench,kernelevolve}. Closed-loop algorithm and program-search systems such as AlphaTensor, AlphaEvolve, OpenEvolve, LoongFlow, ShinkaEvolve, and AVO further motivate generate-evaluate-select loops with evolutionary memory~\citep{alphatensor,alphaevolve,openevolve,loongflow,shinkaevolve,avo}. GPU kernel optimization, however, remains a constrained systems problem: a candidate must satisfy the contest packaging contract, compile in the target container, preserve layout and numerical semantics, and improve mean latency over the full workload distribution rather than a single representative case. Failures therefore often arise from the surrounding engineering loop: stale baselines, incomplete workload coverage, packaging drift, noisy promotion, and loss of profiler or provenance information.

This paper studies harness engineering~\citep{openai_harness,anthropic_harness} as the mechanism for making LLM-driven kernel generation reliable. Prior agent work shows that interface design, interleaved reasoning and action, and verbal feedback memory can materially affect coding-agent behavior~\citep{sweagent,react,reflexion}. In this setting, humans define objectives, resources, feedback loops, and promotion rules, while coding agents perform bounded implementation search. The paper evaluates this design across all FlashInfer contest tracks using official-aligned B200 artifacts. The transferable claim is not that a particular model autonomously solves kernel optimization, but that profile-backed controller state, workload-grounded gates, and artifact memory make LLM search auditable enough to improve real contest kernels.

Our approach instantiates this claim through several key designs that make agent-assisted kernel optimization reproducible rather than prompt-only trial and error.

\noindent\textbf{Skill-Grounded Optimization Harness.}
We encode expert optimization practice as reusable skills rather than one-off prompts. The generic CUDA skill defines the candidate contract, evaluation, profiling, and two-stage search, while the FlashInfer B200 skill adds reference-first reconnaissance, workload discovery, paired gates, official-aligned evaluation, and latency-first promotion.

\noindent\textbf{Profile-Backed Optimization Controller.}
Beyond measuring candidate code, the workflow separates the evaluation harness from the optimization controller. The controller converts NCU and Torch Profiler evidence into bottleneck state, selects one bounded optimization direction, supervises plateaus or regressions, and records accepted and rejected evidence for later rounds.

\noindent\textbf{Workload-Grounded Shape Dispatch.}
The system derives optimization regimes from contest workload UUIDs, JSON workload axes, and measured latency distributions rather than from a single hand-picked input. Shape-specialized routes are introduced only when profile and latency evidence show distinct limiting factors.

\noindent\textbf{Human-Guided Plateau Recovery.}
Human effort remains central: humans design the harness, curate references, decide evaluation budget, and make final promotion decisions. When agents plateau, humans steer the search by switching GPT and Claude model families, requesting sub-agent review, supplying reference kernels or documentation, and changing implementation languages.

\noindent\textbf{Hardware-Aware Language Selection.}
The workflow treats CUDA C++, Triton~\citep{triton}, and CuTe/CUTLASS~\citep{cutlass} as alternative control surfaces, with DeepGEMM used as FP8 GEMM reference material~\citep{deepgemm}. Triton supports rapid specialization, CUDA C++ exposes low-level launch and memory control, and CuTe/CUTLASS provides explicit tensor-core layouts and Blackwell-specific kernels.

\noindent\textbf{Noise-Resistant Promotion and Artifact Memory.}
Candidates are promoted by all-workload latency evidence, paired baselines, correctness pass rate, and repeated validation where needed rather than by an isolated speedup. Rejected probes, profiler reports, shape matrices, and promotion decisions are archived as negative or positive trajectory memory, reducing repeated exploration of failed routes.

\begin{figure*}[!t]
\centering
\includegraphics[width=0.985\textwidth]{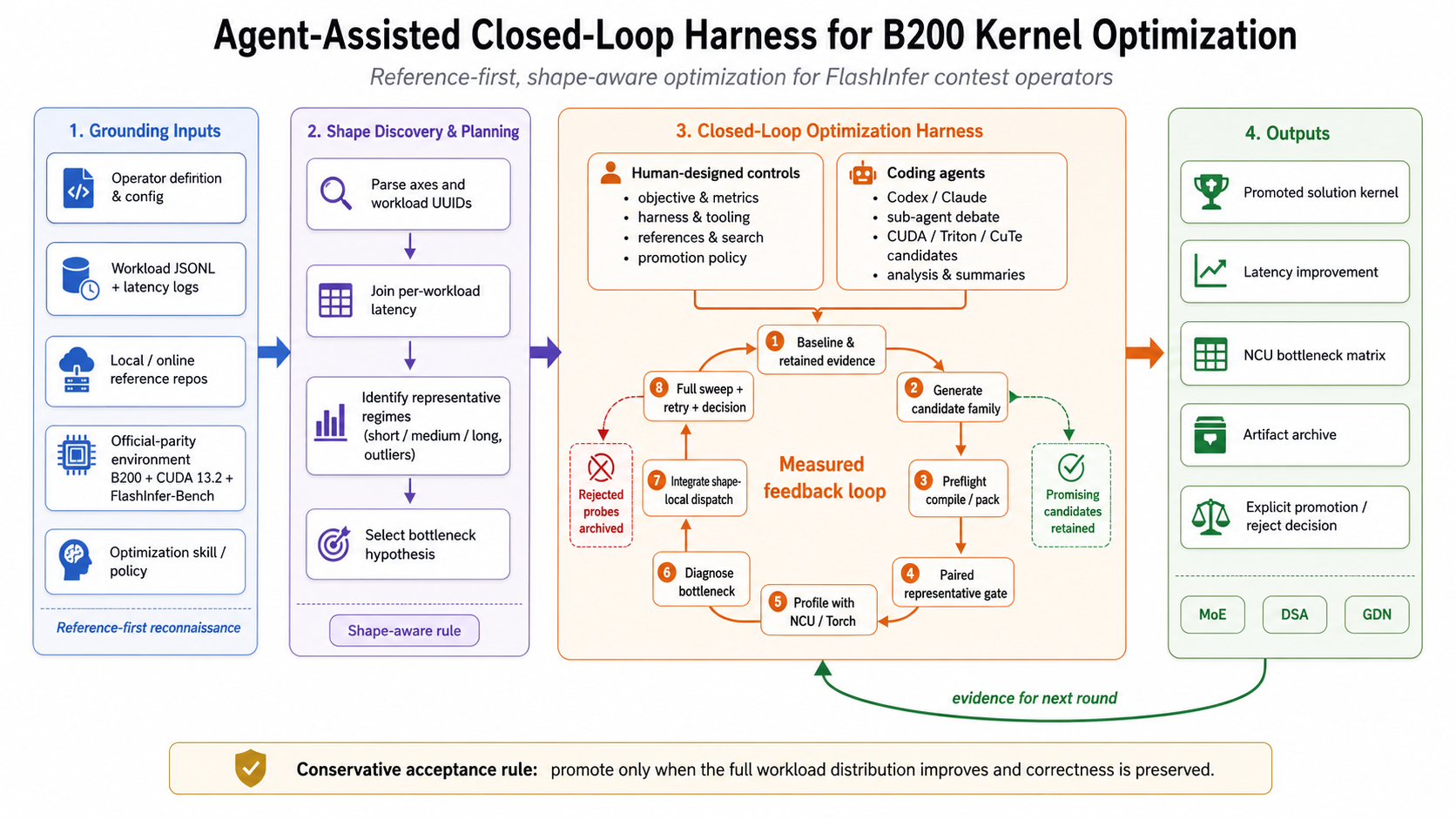}
\caption{Closed-loop harness/controller workflow used for CUDA kernel optimization. The harness measures, archives, and promotes candidates, while the controller structures prompt construction, candidate generation, profiling feedback, and trajectory memory.}
\label{fig:closedloop}
\end{figure*}

The paper covers all three contest tracks and five definitions: FP8 Fused MoE, DSA, and GDN.

\begin{table}[!htbp]
\centering
\caption{Contest definitions covered by this Agent-Assisted paper. Workload denotes the number of official-aligned workloads evaluated for each definition.}
\label{tab:defs}
\setlength{\tabcolsep}{3pt}
\begin{tabularx}{\columnwidth}{@{}lrY@{}}
\toprule
\textbf{Definition} & \textbf{Workload} & \textbf{Kernel stack} \\
\midrule
MoE FP8 & 19 & Triton/CUDA runtime \\
DSA top-k & 128 & CUDA/CuTe via TVM-FFI \\
DSA attention & 23 & Triton dispatcher/CUDA \\
GDN decode & 54 & Triton and CuTe paths \\
GDN prefill & 100 & CuTe Blackwell chunk kernel \\
\bottomrule
\end{tabularx}
\end{table}

The definitions are summarized in Table~\ref{tab:defs}. The retained runs used an official-aligned environment consistent with FlashInfer-Bench: CUDA 13.2, PyTorch 2.12, Triton 3.6, cupti-python timing, isolated runners, and B200 GPUs. Correctness and latency were both mandatory. Latency was the promotion metric; PyTorch reference latency was used as supporting context. This paper does not propose a new serving runtime, model architecture, or autonomous optimizer; it studies the agent-assisted harness and controller needed to generate correct and fast drop-in contest kernels.

\section{Harness System Design}

OpenAI describes harness engineering as a shift from humans writing every line of code to humans designing environments, constraints, and feedback loops that let agents do reliable work~\citep{openai_harness}. Anthropic makes a similar argument for long-running application development harnesses~\citep{anthropic_harness}. In this project, the environment was a CUDA kernel optimization loop rather than an application repository. The harness made the task legible to agents by turning workloads, profiles, rejected variants, and retained baselines into structured context. The architecture used the closed loop in Figure~\ref{fig:closedloop}.

\noindent The loop has four main responsibilities: grounding candidates in operator definitions, reference code, workload JSON files, baselines, and target-environment constraints; discovering shape axes such as sequence length, batch size, page count, and outlier regimes; closing the baseline-profile-generate-evaluate-archive feedback cycle; and producing an explicit promotion decision with latency evidence and profiler artifacts.

\noindent\textbf{Evaluation harness vs. optimization controller.}
The evaluation harness is the measurement layer: it packages candidates, compiles them in the target environment, runs correctness checks, measures official-aligned B200 latency, and stores benchmark and profiler artifacts. The optimization controller is the decision layer: it converts profile evidence into a reusable state, selects the next optimization hypothesis, carries supervisor constraints across rounds, and writes outcomes into memory. The unified CUDA optimization skill codifies this controller role, while the FlashInfer contest harness provides workload, environment, and promotion discipline.

The harness used conservative acceptance rules. Representative gates first compared a candidate against the same-round baseline on selected workloads drawn from the measured workload axes. Full sweeps then evaluated the entire distribution. For large definitions such as DSA top-k and GDN prefill, the multi-workload runner launched one evaluation worker per workload up to a worker limit, wrote incremental JSON, and retried transient infrastructure failures. A candidate was promoted only if the full distribution improved without correctness regressions. Rejected probes were archived because negative evidence prevents repeated exploration of failed launch-bound, fusion, or tile-size hypotheses.

\section{Optimization Workflow}

\subsection{Agent and Prompting Workflow}

The optimization workflow used Codex and Claude Code as coding agents. The primary model families were GPT-5 variants, especially GPT-5.3-Codex~\citep{gpt53codex}, together with Claude Opus 4.6~\citep{claudeopus46} for independent reasoning and code review. Humans supplied the harness, contest-specific references, evaluation policy, and final promotion criteria. Agents generated candidate patches, diagnosed profiles, proposed route changes, and summarized failures.

Skills were used as reusable execution playbooks. Codex skills package instructions, references, and optional scripts behind \code{SKILL.md}, while Claude Code skills provide analogous project or personal workflows~\citep{codex_skills,claude_skills}. The harness is described through three complementary skill layers. The generic CUDA kernel optimizer standardized the Model/ModelNew contract, correctness and performance evaluation, NCU/NSYS profiling, two-stage search, and export of reusable candidates. The FlashInfer B200 contest optimizer specialized this loop to reference-first search, real workload-shape discovery, same-round paired gates, B200/CUDA 13.2 tactics, official-parity evaluation, and artifact archival. The unified CUDA optimization controller codifies a profile-backed round contract: state extraction, state matching, one-direction optimization selection, candidate generation, trajectory supervision, knowledge-base update, and manifest-based resumption. This reduced prompt drift and made repeated optimization rounds reproducible across operators, model sessions, and implementation languages.

Sub-agents were used for parallel exploration, debate, and review~\citep{codex_subagents}. In practice, this enabled independent review of shape regimes, profiler interpretations, and candidate routes when the primary agent stopped finding measured improvements. The harness still kept promotion centralized, so sub-agent outputs were treated as proposals rather than authority.

The main workflow studied in this paper is Agent-Assisted: humans author the harness, curate references, steer plateau recovery, and approve promotions under measured gates. We also ran Full-Agent experiments with LoongFlow PES as same-protocol autonomous-search baselines~\citep{loongflow}. Their matched final evaluations are reported in Section~\ref{sec:experimental-results}; Appendix~\ref{sec:appendix-full-agent} summarizes the search traces that produced those artifacts.

\begin{algorithm}[t]
\caption{Agent-Assisted Harness Loop}
\label{alg:agent-assisted-loop}
\small
\begin{algorithmic}
   \STATE {\bfseries Input:} definition $D$, workloads $W$, languages $L$, hardware $H$
   \STATE \hspace{1.6em} FlashInfer baseline $B$, round budget $N$, promotion gate $G$
   \STATE {\bfseries Output:} retained solution $\mathrm{Sol}^{*}$ and artifact archive $\mathcal{A}$
   \STATE $\mathcal{A} \leftarrow \emptyset$
   \STATE $\mathcal{M} \leftarrow \mathrm{InitializeMemory}(D,W,L,H,B)$
   \STATE $\mathrm{Sol}_{0} \leftarrow B$
   \FOR{$i \leftarrow 0$ {\bfseries to} $N-1$}
      \STATE $\Delta_i \leftarrow \emptyset$
      \STATE $s_i \leftarrow \mathrm{Controller.State}(\mathcal{M}, \mathrm{Sol}_{i})$
      \STATE $C_i \leftarrow \mathrm{Agent.Generate}(\mathrm{Sol}_{i}, s_i, L)$
      \FOR{{\bfseries each} candidate $c \in C_i$}
         \STATE $r \leftarrow \mathrm{Harness.RepGate}(D,W,c,\mathrm{Sol}_{i},H)$
         \IF{$r.\mathrm{status}=\mathrm{PASSED}$}
            \STATE $p \leftarrow \mathrm{Profiler.Summarize}(c,r)$
            \STATE $\mathcal{A} \leftarrow \mathcal{A} \cup \{(c,r,p)\}$
         \ENDIF
      \ENDFOR
      \STATE $c_i^{+} \leftarrow \mathrm{SelectBestPassed}(\mathcal{A}, i)$
      \IF{$c_i^{+}$ exists}
         \STATE $R_i \leftarrow \mathrm{Harness.FullSweep}(D,W,c_i^{+},H)$
         \STATE $\Delta_i \leftarrow R_i$
         \IF{$\mathrm{Promote}(R_i,\mathrm{Sol}_{i},G)$}
            \STATE $\mathrm{Sol}_{i+1} \leftarrow \mathrm{IntegrateOrDispatch}(\mathrm{Sol}_{i},c_i^{+},R_i)$
         \ELSE
            \STATE $\mathrm{Sol}_{i+1} \leftarrow \mathrm{Sol}_{i}$
         \ENDIF
      \ELSE
         \STATE $\mathrm{Sol}_{i+1} \leftarrow \mathrm{Sol}_{i}$
      \ENDIF
      \STATE $\mathcal{M} \leftarrow \mathrm{UpdateMemory}(\mathcal{M},\mathcal{A},\Delta_i)$
      \IF{$\mathrm{Supervisor.Plateau}(\mathcal{M})$}
         \STATE $\mathcal{M} \leftarrow \mathrm{HumanSteer}(\mathcal{M})$
      \ENDIF
   \ENDFOR
   \STATE $\mathrm{Sol}^{*} \leftarrow \mathrm{BestCorrectFullSweep}(\mathcal{A})$
   \STATE {\bfseries return} $\mathrm{Sol}^{*}, \mathcal{A}$
\end{algorithmic}
\end{algorithm}

Algorithm~\ref{alg:agent-assisted-loop} summarizes the loop. Agents propose candidates; the harness handles compilation, correctness, profiling, full sweeps, and promotion. The controller is human-authored evidence memory, not autonomous; plateau steering updates context without bypassing measured gates.

\subsection{Human vs. Agent Contributions}

The collaboration followed an AlphaEvolve closed loop: humans defined the objective, constraints, evaluation surface, and acceptance policy, while agents searched within that engineered environment~\citep{alphaevolve}. The primary objective was to minimize average latency over all contest shapes; reported speedups use the normalizations stated in Section~\ref{sec:experimental-results}. Secondary signals included per-shape regressions, median latency, 95th-percentile (P95) latency, and outlier regimes that required separate dispatch.

Human effort focused on the design and orchestration of the optimization harness. We designed CUDA kernel optimization skills and developed serial and parallel B200 evaluation scaffolds. We integrated Torch Profiler and NVIDIA Nsight Compute (NCU) analysis scripts, and curated documentation and local references from FlashInfer, DeepGEMM, TensorRT-LLM, and contest-relevant GPU inference repositories~\citep{deepgemm,tensorrtllm}. To address gaps in local code, we directed web and GitHub reference searches. To overcome optimization plateaus, we switched GPT and Claude model families, changed implementation surfaces, and used sub-agent debate. We also designed shape-aware dispatch strategies, but only promoted dispatch routes when profiler and latency evidence showed that the relevant shape regime had a distinct limiter.

Agent work centered on implementation search under those controls. Codex and Claude Code generated CUDA, Triton, and CuTe candidates; produced multiple parallel samples across different optimization directions; adapted reference kernels; proposed route and dispatch changes; interpreted profiler output; and suggested follow-up experiments. Candidate code was not accepted by assertion: it had to pass harness-controlled correctness, compilation, profiling, and latency gates. Appendix~\ref{sec:human-agent-appendix} summarizes the contribution split.

\subsection{Iterative Refinement Loop}

Each round instantiated Algorithm~\ref{alg:agent-assisted-loop} with a fixed sequence: identify the bottleneck and workload regime from retained runs and NCU data; generate a bounded candidate family around one hypothesis; pack and preflight-compile the candidate; run a paired representative gate against the same-round baseline; profile correct and promising candidates; integrate the best candidate behind a cheap dispatch rule if the win is shape-local; run a full sweep and retry transient infrastructure failures; and promote only if the full distribution improves without correctness regressions.

\noindent\textbf{Profile-backed state and trajectory control.}
The unified controller codifies the stateful version of the loop. A closed round can produce a profiler-derived state signature, a state-match decision, one selected optimization, a candidate and evaluation result, a supervisor decision, and a knowledge-base update. The supervisor detects stalling, cycles, regressions, plateaus, and diminishing returns, then carries blocked or recommended directions into the next round. This loop follows the same structure as closed-loop program search: the model proposes code, the evaluator returns a scalar signal and diagnostics, and the system retains only measured improvements~\citep{alphaevolve}. The contest-specific difference is that the evaluator must also enforce packaging, correctness, target-environment compilation, and workload-distribution coverage.

\begin{table}[t]
\centering
\caption{Implementation artifacts referenced by this paper. These repositories are reproducibility artifacts, not official leaderboard claims.}
\label{tab:artifacts}
\setlength{\tabcolsep}{3pt}
\begin{tabularx}{\columnwidth}{lY}
\toprule
\textbf{Track} & \textbf{Repository artifact} \\
\midrule
All tracks & \href{https://github.com/syhya/mlsys26-flashinfer-contest}{mlsys26-flashinfer-contest}~\citep{contestartifact} \\
Fused MoE & \href{https://github.com/syhya/mlsys26-flashinfer-solution-fused-moe}{mlsys26-flashinfer-solution-fused-moe}~\citep{moeartifact} \\
Sparse Attention & \href{https://github.com/syhya/mlsys26-flashinfer-solution-sparse-attention}{mlsys26-flashinfer-solution-sparse-attention}~\citep{sparseartifact} \\
Gated Delta Net & \href{https://github.com/syhya/mlsys26-flashinfer-solution-gated-delta-net}{mlsys26-flashinfer-solution-gated-delta-net}~\citep{gdnartifact} \\
\bottomrule
\end{tabularx}
\end{table}

\section{Experimental Evaluation}
\label{sec:experimental-results}

The primary results in this paper are the Agent-Assisted retained official-aligned B200 measurements from archived artifacts. They are not undisclosed final official leaderboard scores. We also built Full-Agent implementations using LoongFlow PES and evaluated the selected Full-Agent artifacts on the same official-aligned workload sets, evaluation protocol, and supplied FlashInfer baselines used for the Agent-Assisted artifacts~\citep{loongflow}. Following the efficiency-table style used in LoongFlow, Table~\ref{tab:results} reports common mean-latency speedup normalizations. The PyTorch column uses the PyTorch reference mean recorded in the corresponding Agent-Assisted retained evaluator artifact, while the FlashInfer column is normalized to the supplied FlashInfer baseline; higher speedup and lower latency are better. The DSA top-k entry preserves the low-trial retained artifact for comparability, while Appendix~\ref{sec:appendix-dsa-topk-repeated-validation} separately records the high-trial replay evidence and conservative fallback tag.

\begin{table*}[t]
\centering
\caption{Efficiency comparison sorted by retained FlashInfer-relative mean-latency speedup. Latencies are mean milliseconds; lower is better. PyTorch speedup uses the retained Agent-Assisted PyTorch reference mean for each definition; FlashInfer speedup uses the supplied FlashInfer baseline. Full-Agent rows are matched final-evaluation baselines evaluated on the same official-aligned workload sets.}
\label{tab:results}
\small
\setlength{\tabcolsep}{4pt}
\begin{tabular}{@{}llrrr@{}}
\toprule
\textbf{Definition} &
\textbf{Method} &
\textbf{\shortstack{Latency\\(ms)}} &
\textbf{\shortstack{Speedup\\vs. PyTorch}} &
\textbf{\shortstack{Speedup\\vs. FI}} \\
\midrule
\textbf{DSA Attention} & \textbf{Agent-Assisted} & \textbf{0.011175} & \textbf{217.17$\times$} & \textbf{29.68$\times$} \\
 & Full-Agent & 0.022811 & 106.39$\times$ & 14.54$\times$ \\
 & FlashInfer baseline & 0.331650 & 7.32$\times$ & 1.00$\times$ \\
\midrule
\textbf{DSA Indexer} & \textbf{Agent-Assisted} & \textbf{0.006893} & \textbf{494.13$\times$} & \textbf{18.05$\times$} \\
 & Full-Agent & 0.032659 & 104.29$\times$ & 3.81$\times$ \\
 & FlashInfer baseline & 0.124420 & 27.38$\times$ & 1.00$\times$ \\
\midrule
\textbf{GDN Prefill} & \textbf{Agent-Assisted} & \textbf{0.051992} & \textbf{21{,}078$\times$} & \textbf{13.70$\times$} \\
 & Full-Agent & 0.688875 & 1{,}591$\times$ & 1.03$\times$ \\
 & FlashInfer baseline & 0.712166 & 1{,}539$\times$ & 1.00$\times$ \\
\midrule
\textbf{MoE FP8} & \textbf{Agent-Assisted} & \textbf{0.28634} & \textbf{63.78$\times$} & \textbf{1.62$\times$} \\
 & FlashInfer baseline & 0.463874 & 39.37$\times$ & 1.00$\times$ \\
 & Full-Agent & 1.74263 & 10.48$\times$ & 0.27$\times$ \\
\midrule
\textbf{GDN Decode} & \textbf{Agent-Assisted} & \textbf{0.006201} & \textbf{7{,}970$\times$} & \textbf{1.12$\times$} \\
 & FlashInfer baseline & 0.006940 & 7{,}121$\times$ & 1.00$\times$ \\
 & Full-Agent & 0.008366 & 5{,}907$\times$ & 0.83$\times$ \\
\bottomrule
\end{tabular}
\end{table*}

\textbf{Under the matched final-evaluation setting, the Agent-Assisted retained kernels are faster than the Full-Agent artifacts.} The selected Full-Agent artifacts are \textbf{1.35--13.25$\times$ slower} than the Agent-Assisted retained artifacts under the same FlashInfer-relative normalization, and two remain below the supplied FlashInfer baseline: MoE FP8 at \textbf{0.27$\times$} and GDN Decode at \textbf{0.83$\times$}. Section~\ref{sec:final-kernel-analysis} analyzes the retained \textbf{Agent-Assisted kernels}, whose stronger results depended on human harness design, reference selection, and conservative promotion gates.

\begin{figure}[t]
\centering
\includegraphics[width=0.74\linewidth]{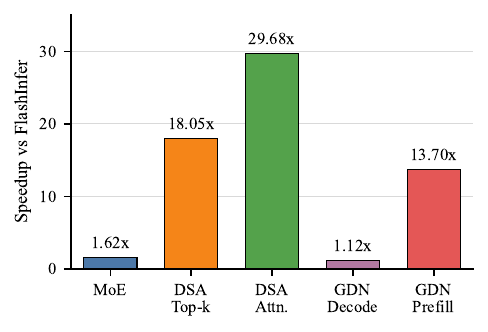}
\caption{Final retained mean-latency speedup over the supplied FlashInfer baseline.}
\label{fig:baseline}
\end{figure}

Following the FlashInfer-Bench evaluation protocol~\citep{flashinferbench,contest}, speedup is correctness-gated and computed over workload-level latency ratios. For a definition $d$ with workload set $W_d$, FlashInfer baseline latency $b_{d,w}$, retained kernel latency $\ell_{d,w}$, and definition-level correctness indicator $c_d \in \{0,1\}$, the official per-kernel and per-track scores are
\[
S_d = c_d \cdot \frac{1}{|W_d|}\sum_{w \in W_d}\frac{b_{d,w}}{\ell_{d,w}},
\qquad
S_t = \frac{1}{E_t}\sum_{d \in D_t} S_d,
\]
where $E_{\mathrm{MoE}}=1$, $E_{\mathrm{DSA}}=2$, and $E_{\mathrm{GDN}}=2$. If any workload fails correctness, then $c_d=0$; a missing definition in DSA or GDN contributes zero and effectively halves a single-definition submission. Table~\ref{tab:results} and Figure~\ref{fig:baseline} instead use a simpler ratio-of-means latency summary to compare FlashInfer, Full-Agent, and Agent-Assisted methods; this is a reporting normalization, not the official contest score. With mean FlashInfer baseline latency $\bar{b}_d$ and mean retained latency $\bar{\ell}_d$, the reported speedup is
\[
\mathrm{Speedup}_d =
\frac{\text{mean FlashInfer baseline latency}}
     {\text{mean retained solution latency}}
= \frac{\bar{b}_d}{\bar{\ell}_d}.
\]
For example, the MoE summary speedup is $0.463874 / 0.286342 = 1.62\times$. Higher values indicate lower average latency relative to the supplied FlashInfer baseline.

Figure~\ref{fig:baseline} visualizes the final mean-latency speedup of each retained operator over the supplied FlashInfer baseline. The largest FlashInfer-baseline speedups occurred on sparse attention and DSA top-k, where structural route changes reduced work at dominant regimes. GDN prefill also showed a large speedup by replacing the main algorithmic path. GDN decode had the smallest speedup because the supplied FlashInfer baseline was already close to the retained route.

\begin{figure*}[!t]
\centering
\includegraphics[width=0.965\textwidth]{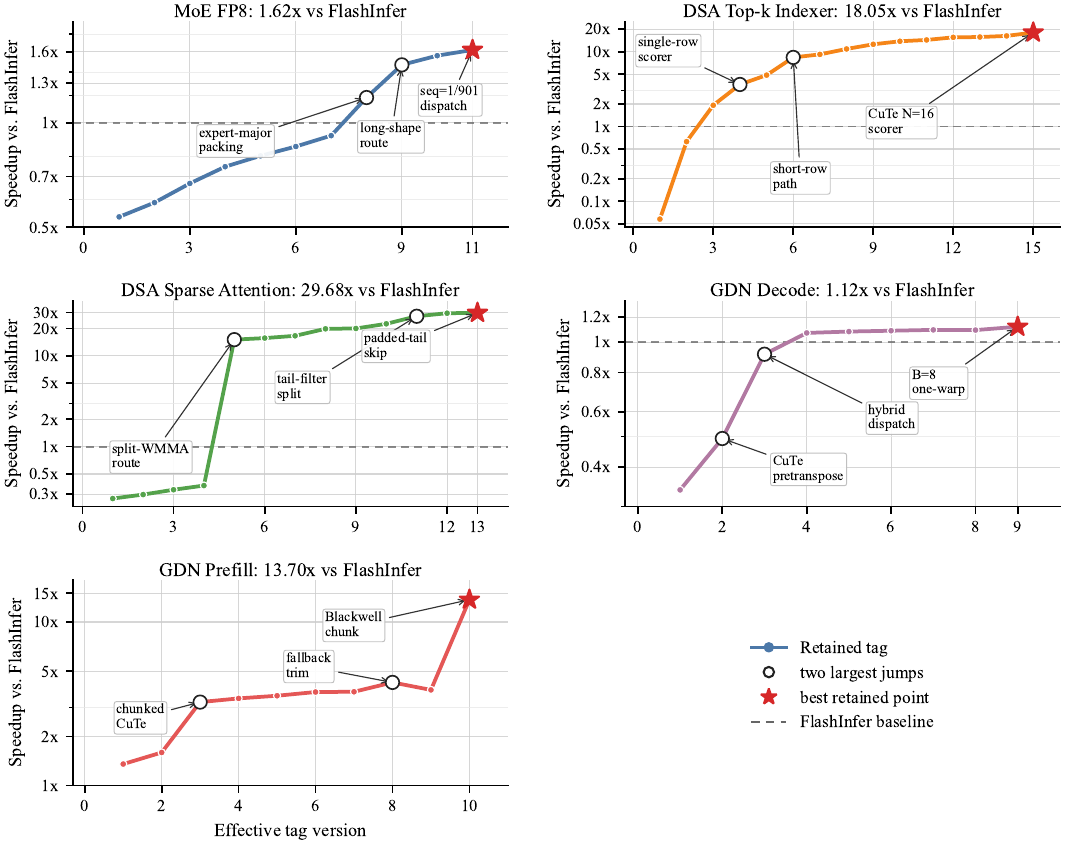}
\caption{Retained speedup trajectories over the supplied FlashInfer baseline. The y-axis is log-scale speedup versus FlashInfer; the x-axis uses effective tag versions after collapsing retained tags with unchanged latency. Open circles mark the two largest non-final jumps, stars mark the best retained points, and dashed lines mark the \(1.0\times\) FlashInfer baseline.}
\label{fig:trajectory}
\end{figure*}

\section{Final Kernel Analysis}
\label{sec:final-kernel-analysis}

\subsection{Per-Operator Case Studies}

Having established the retained results, this section analyzes the final kernels rather than the search chronology. The final kernels were not produced by a single universal optimization. The effective pattern was shape-aware dispatch combined with full-workload promotion: each candidate route was retained only when it improved the measured distribution without breaking correctness. This policy is important because the contest workloads mix launch-bound regimes, tensor-core regimes, irregular sparse-memory regimes, and recurrent state updates.

For MoE FP8, the bottleneck is the transition from top-k routed tokens to expert-major matrix multiplications and then back to token order. The selected design uses a CUDA helper to fuse routing, local-expert filtering, counting, prefix construction, and FP8 packing into a compact workspace. Triton persistent grouped general matrix multiply (GEMM) kernels then execute the two expert matrix products while handling block-scale dequantization and stable accumulation. The retained variant kept the expert-major mainline, used a shape-local epilogue path for the single-token regime, and preserved a routing and packing specialization for the long but structured sequence-length regime. Forced rerouting of additional shapes was rejected because it improved some local cases but regressed the full sweep~\citep{triton,deepgemm,moeartifact}.

For the DSA top-k indexer, the optimization separates score generation from top-k selection. The scorer is a CUDA/CuTe tensor-core kernel over FP8 query and key inputs, with vectorized memory access and a tile shape chosen to reduce launch count while improving query reuse in the medium-band workloads. The selector preserves the top-k semantics through three regimes: a vectorized filtered path for common cases, a histogram fallback for wider candidate ranges, and a short-row pass-through path when the scoring work is already small. This separation made it possible to tune the high-throughput scoring path without destabilizing correctness-sensitive selection logic~\citep{sparseartifact}.

\textbf{Trial-sensitive correctness.} A post-submission DSA top-k check exposed a trial-sensitive numerical issue that was not consistently surfaced by the low-trial validation used during fast agent iterations. The source of this inconsistency is that FlashInfer-Bench evaluates randomized per-trial inputs and states, while the resolved default budget used by the harness is only \(n=3\) trials~\citep{flashinferbench}; rare top-k boundary cases can therefore be missed by one low-trial sweep and exposed by another. This is a practical limitation of kernel-search harnesses: repeated high-trial evaluation over every candidate and workload is often too expensive on B200, so the loop uses representative gates and low-trial full sweeps to preserve iteration speed. The follow-up evidence in Appendix~\ref{sec:appendix-dsa-topk-repeated-validation} shows that targeted high-trial replay can expose rare selector-boundary failures and identify a more conservative fallback tag. We therefore treat high-trial replay as a final validation gate for suspicious or high-impact DSA top-k shapes rather than as the default inner-loop gate.

For DSA sparse attention, the principal challenge is that the sparse index set induces different bottlenecks depending on token count and page-table representability. Short decode workloads are best served by a small specialized route or an upstream-derived helper route. Larger workloads use split key-value (KV) flash-decoding with asynchronous sparse key gathers, shared-memory layouts chosen to reduce bank conflicts, and a separate merge phase for partial outputs. The accepted final change targets padded tail work in the large split route: it avoids unnecessary attention and value accumulation on empty upper-half tiles while keeping the same output contract for downstream merging~\citep{pagedattention,sparseartifact}.

For GDN decode, the retained implementation is explicitly dispatch-driven. A single recurrent kernel did not generalize across batch regimes, because small batches are dominated by launch and state-access overhead while larger batches need different pretranspose and value-tile behavior. The final dispatcher therefore combines a Triton recurrent route for small batches, a one-warp specialization for the batch-eight regime, a CuTe pretranspose route for mid-sized batches, and a larger-value route for the highest measured batch regime. This explains why GDN decode shows the smallest final speedup: the supplied baseline was already strong, and the remaining wins were batch-local rather than global~\citep{gdnartifact}.

For GDN prefill, the key improvement is algorithmic routing rather than a narrow micro-optimization. The final implementation makes a Blackwell chunked CuTe-DSL kernel the default path, replacing a broader dispatch table that was difficult to tune consistently. The chunked path uses SM100 tensor memory, Tensor Memory Accelerator (TMA) movement, and warp specialization to separate matrix products, gate loading, recurrence correction, and output storage. A scalar preprocessing helper prepares the gate and beta inputs for the chunked kernel, while a narrow recurrent fallback handles tiny shapes where the chunked path is not yet favorable~\citep{gdnartifact}.

\subsection{Implementation Stack}

\noindent\textbf{Language selection as part of the search policy.}
The implementation used multiple authoring surfaces because each operator exposed a different limiting factor. Triton was most effective where rapid specialization and maintainable grouped matrix kernels mattered, as in MoE and the small-batch GDN decode path~\citep{triton}. CUDA C++ was used where direct control over shared memory, launch attributes, vectorized memory access, and low-level conversion instructions determined performance, especially in DSA top-k and DSA sparse attention. CuTe and CUTLASS provided explicit tensor-core layouts for DSA top-k and the Blackwell chunked GDN prefill path~\citep{cutlass}. Python remained the orchestration layer for dispatch, packaging, caching, and harness-controlled exclusion of variants that were correct but not globally beneficial. Language switching was therefore a controlled response to profiler evidence and implementation constraints, not an aesthetic preference.

\subsection{Profile-Guided Analysis}

\noindent\textbf{Profiler output as decision evidence.}
Torch Profiler and NCU outputs were compressed into decision records: dominant kernel, workload regime, bottleneck class, limiting resource, occupancy, throughput, waves per SM, and candidate headroom. This made profiling actionable for dispatch and promotion. A route could win a local UUID or shape band and still be rejected when the full sweep showed that its benefit did not survive workload mixing.

The condensed largest-shape evidence used for these decisions is reported in Appendix~\ref{sec:appendix-profile-evidence}, especially Table~\ref{tab:appendix-profile-summary}. Across the five operators, the profiler evidence supports the same policy: keep route changes only when Torch Profiler and NCU identify a distinct limiting regime, such as GEMM-plus-packing cost in MoE, low-wave sparse launches in DSA, and resource pressure in the GDN prefill chunk kernel.

\section{Limitations and Future Directions}

The main limitation is that this paper evaluates an Agent-Assisted engineering workflow, not an isolated LLM. Human harness design, budget decisions, reference selection, and final promotion were central. The workload distributions are contest-specific, and the reported timings remain local official-aligned B200 evidence rather than final leaderboard scoring.

Future work should move more controller state, memory, and model orchestration into the harness while preserving measured promotion discipline. Longer-running research agents and kernel-specific systems suggest how to increase candidate diversity~\citep{autoresearch,autokernel,kernelgensurvey,kernelevolve}. Evolutionary memory systems and agentic variation operators offer another route for broadening candidate search~\citep{alphaevolve,openevolve,loongflow,shinkaevolve,avo}, but the DSA top-k replay shows that stronger automation must also strengthen final validation gates.

\section{Conclusion}
\label{sec:main-end}

This paper presented an Agent-Assisted workflow for FlashInfer contest kernel optimization on NVIDIA B200 GPUs. The central result is a measurable systems loop rather than a prompt template: reusable CUDA optimization skills, a contest-specific evaluation harness, a profile-backed controller, workload-grounded shape dispatch, and artifact memory convert agent output into compile, correctness, profiling, full-sweep, and promotion decisions. Across five definitions, the retained low-trial official-aligned artifacts achieved FlashInfer-baseline mean-latency speedups from 1.12$\times$ to 29.68$\times$, with the DSA top-k high-trial fallback evidence separated in Appendix~\ref{sec:appendix-dsa-topk-repeated-validation}.

The results show that reliable LLM-driven kernel generation still depends on harness engineering. The largest gains came from structural, hardware-aware changes such as sparse-attention route specialization and the Blackwell chunked GDN prefill path, while matched Full-Agent final evaluations remained slower than the retained Agent-Assisted artifacts. This gap suggests that near-term kernel agents are most useful when paired with human-curated references, profiler interpretation, conservative promotion gates, and persistent trajectory memory; future autonomous systems should move more of this controller and memory into the harness without weakening measured validation.

\section*{Acknowledgment}

We are grateful to Wanping Zhang, Bowen Ren, Zejia Liu, Bo Pang, Yalu Ouyang, and Shiyong Li for helpful discussions, technical feedback, and project support. We also acknowledge the FlashInfer team for organizing the AI Kernel Generation Contest, defining the benchmark tasks, and maintaining the FlashInfer-Bench evaluation infrastructure. Access to NVIDIA B200 GPUs was provided by Modal and enabled the contest-aligned evaluation, profiling, and repeated validation reported in this paper.

\begingroup
\raggedright
\sloppy
\setlength{\bibsep}{5pt plus 1ex}
\clearpage
\bibliographystyle{mlsys2025}
\bibliography{flashinfer_contest_report}
\endgroup

\clearpage
\onecolumn
\raggedbottom
\appendix

\section*{Appendix}
\noindent This appendix provides the Full-Agent trajectory artifacts, human/agent contribution matrix, core skill and prompt excerpts, ablation notes, DSA top-k repeated-validation evidence, operator background, largest-shape NCU/Torch Profiler evidence, and per-workload shape tables used to support the main paper.
\vspace{0.5em}

\section{Full-Agent Trajectory Details}
\label{sec:appendix-full-agent}

In addition to the Agent-Assisted workflow studied in the main text, we also ran Full-Agent experiments with LoongFlow PES~\citep{loongflow}. Table~\ref{tab:appendix-full-agent-summary} summarizes the Full-Agent search traces that produced the selected artifacts. \textbf{The Full-Agent rows in Table~\ref{tab:results} use those selected artifacts evaluated under the matched final-evaluation protocol; the trace-workload counts below describe the search logs, not Table~\ref{tab:results} final-evaluation coverage.} The LoongFlow trace-local best-score checkpoint can differ from the FlashInfer-relative best-latency checkpoint because the trace score is normalized by a run-local PyTorch reference latency.

\begin{table}[!htbp]
\centering
\caption{\normalfont Full-Agent search trace summary. FI speedup is normalized to the supplied FlashInfer baseline; trace workloads describe trace logs rather than Table~\ref{tab:results} final-evaluation coverage.}
\label{tab:appendix-full-agent-summary}
\small
\setlength{\tabcolsep}{4pt}
\begin{tabular}{@{}lrrrrrr@{}}
\toprule
\textbf{Definition} &
\textbf{\shortstack{Trace\\workloads}} &
\textbf{\shortstack{Eval.\\iters}} &
\textbf{\shortstack{FI-best\\iter}} &
\textbf{\shortstack{Latency\\(ms)}} &
\textbf{\shortstack{FI\\speedup}} &
\textbf{\shortstack{Trace-best\\iter}} \\
\midrule
MoE FP8 & 19 & 40 & 40 & 1.44809 & \textbf{0.32$\times$} & 40 \\
DSA Indexer & 12 & 20 & 16 & 0.032659 & \textbf{3.81$\times$} & 14 \\
DSA Attention & 23 & 29 & 29 & 0.022811 & \textbf{14.54$\times$} & 20 \\
GDN Decode & 54 & 10 & 10 & 0.008366 & \textbf{0.83$\times$} & 10 \\
GDN Prefill & 100 & 10 & 9 & 0.688875 & \textbf{1.03$\times$} & 10 \\
\bottomrule
\end{tabular}
\end{table}

\begin{figure}[!htbp]
\centering
\includegraphics[width=0.90\textwidth]{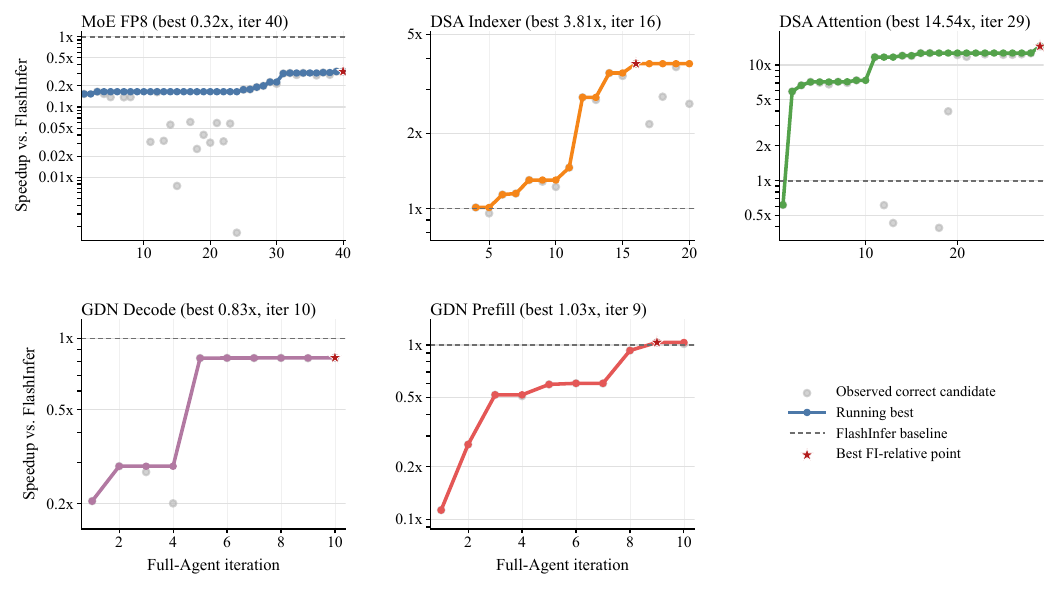}
\caption{\normalfont Full-Agent optimization trajectories extracted from the LoongFlow trace logs. Gray dots are correctness-passing evaluated candidates, solid lines are the running best FlashInfer-relative speedup, dashed lines mark the supplied FlashInfer baseline, and stars mark the best FlashInfer-relative latency point from Table~\ref{tab:appendix-full-agent-summary}. The y-axes are logarithmic because the traces span sub-baseline and multi-\(\times\) regimes.}
\label{fig:appendix-full-agent-trajectory}
\end{figure}

\Needspace{16\baselineskip}
\section{Human and Agent Contribution Matrix}
\label{sec:human-agent-appendix}

\begin{table}[H]
\centering
\caption{Human and agent contributions in the Agent-Assisted workflow.}
\label{tab:human-agent}
\setlength{\tabcolsep}{4pt}
\begin{tabularx}{\textwidth}{lYY}
\toprule
\textbf{Dimension} & \textbf{Human-designed controls} & \textbf{Agent-generated work} \\
\midrule
Objective and metrics & Mean all-shape latency as primary objective; per-shape, median, and P95 latency as secondary checks. & Summarized trade-offs and proposed experiments from measured results. \\
Harness and tooling & Serial/parallel B200 evaluators, profiler scripts, optimization skills, archival, and promotion policy. & Ran bounded candidate iterations and consumed feedback from correctness, profiling, and sweep results. \\
References and search & Curated high-quality references and requested targeted web/GitHub searches. & Adapted CUDA/Triton/CuTe kernels and proposed reusable route changes. \\
Search strategy & Model/language switching, multi-model review, sub-agent debate, parallel sampling, and shape dispatch. & Generated diverse candidate families and refined promising variants under harness gates. \\
\bottomrule
\end{tabularx}
\end{table}

\section{Core Skill and Prompt Template for Iterative Optimization}
\label{sec:appendix-core-skill-prompt}

The core iterative playbook was encoded as a reusable optimization skill rather than an ad hoc prompt. The skill converted each optimization round into a constrained protocol: reference-first reconnaissance, workload-derived shape discovery, paired baseline gates, profiling-guided candidate generation, controller state update, artifact archival, and promotion by repeatable mean-latency improvement. Its purpose was to make agent outputs comparable across operators, model sessions, and implementation languages.

\begin{table}[H]
\centering
\caption{Core skill controls used in the iterative optimization loop.}
\label{tab:appendix-core-skill}
\setlength{\tabcolsep}{4pt}
\begin{tabularx}{\textwidth}{lY}
\toprule
\textbf{Control} & \textbf{Workflow role} \\
\midrule
Reference scan & Inspect local references before candidate work; use external kernels only for gaps. \\
Shape regimes & Parse workload JSON to derive shape axes, heavy regimes, and transition outliers from measured data. \\
Paired gates & Compare each promising candidate against a same-round baseline on representative workloads before a full sweep. \\
Profile matrix & Record per-regime dominant kernels, bottleneck class, resource limiter, occupancy, throughput, and waves per SM. \\
State controller & Convert successful profiles into state signatures, match them to prior evidence, select one optimization direction, and preserve manifest state. \\
Supervisor memory & Detect stalls, cycles, regressions, plateaus, and diminishing returns; carry blocked or recommended directions into later rounds. \\
B200 tactics & Treat clusters, DSMEM, L2 persistence, shared-memory carveout, and dependent launch as evidence-driven branches. \\
Explore/exploit & Explore architecture-diverse candidates first, then exploit the top routes with tile, launch, register, occupancy, and dispatch tuning. \\
Archive/promote & Archive kernels, benchmark JSON, profiler summaries, shape matrices, and decisions; promote only if the full distribution improves without correctness regressions. \\
\bottomrule
\end{tabularx}
\end{table}

The full skill is longer than a paper appendix, so the excerpt below records the prompt-style core that was repeatedly exposed to the coding agent. This follows the artifact style of prompt appendices: it specifies the role, required inputs, hard gates, step order, and final decision contract rather than only describing the method in prose.

\Needspace{20\baselineskip}
\subsection{FlashInfer B200 Optimization Skill Excerpt}
\label{sec:appendix-skill-excerpt}

\begingroup
\small
\setlength{\parskip}{2pt}
\noindent\rule{\textwidth}{0.4pt}

\textbf{Role.} Optimize one FlashInfer contest definition for NVIDIA B200 / CUDA 13.2. Treat the official FlashInfer-Bench evaluator, isolated runner, correctness checks, and measured latency as the ground truth.

\textbf{Round input.} Active definition from \code{config.toml}; current retained mean latency; workload JSONL with UUIDs and axes; paired same-round baseline measurements; compact NCU/Torch Profiler bottleneck table; permitted implementation surface: Triton, CUDA C++, CuTe/CUTLASS, or Python dispatch.

\textbf{Protocol.} Run Stage 0 reference-first reconnaissance before candidate code; derive shape regimes from the actual contest workload file by joining UUIDs, axes, and retained latency; gate early candidates on two or three representative regimes with a same-round baseline; profile promising routes into compact bottleneck evidence; branch CUDA 13.2 / Blackwell tactics only when a measured limiter supports them; and promote only after official-parity full-sweep evidence with every workload passing.

\textbf{Hard gates.} Do not start candidate writing before reference reconnaissance; do not hardcode shape thresholds from another operator; do not promote a single-workload win as a global result; do not keep a B200-only feature that wins one outlier but loses the representative mean; do not report an incomplete sweep as promotion evidence.

\textbf{Decision output.} End each round with exactly one decision: archive only, promote globally, or reject and restore. Attach the benchmark JSON, profiler summary, candidate kernel, workload matrix, and reason for the decision.

\noindent\rule{\textwidth}{0.4pt}
\endgroup

The corresponding prompt template exposed the same controls to the coding agent. Each round specified: the active operator, correctness contract, packaging constraints, and current best latency; the reference and reconnaissance summary; the workload-derived regimes, paired baseline measurements, and latency-heavy outliers; the profiler bottleneck table and optimization hypothesis for the next candidate family; the permitted implementation surfaces, such as Triton, CUDA C++, or CuTe; the evaluation plan, including representative gates, repeat validation, optional full sweep, retry policy, and promotion threshold; and the archival requirement with a final decision label of archive only, promote globally, or reject and restore.

A typical instruction therefore asked the agent to generate a bounded candidate family around one bottleneck, run or prepare paired gates, interpret profiler output, and update the retained route only through the harness decision policy. This prompt design prevented a single-shape local win from being treated as a submission-ready result.

\Needspace{20\baselineskip}
\section{Ablation Notes}

Table~\ref{tab:appendix-ablation} records representative ablations that changed promotion decisions. Each row is backed by retained summaries or rejected probes; the purpose is to explain why a route was kept or rejected, rather than to enumerate every trial.

\begin{table}[H]
\centering
\caption{Representative ablation evidence from retained summaries and rejected probes.}
\label{tab:appendix-ablation}
\setlength{\tabcolsep}{4pt}
\begin{tabularx}{\textwidth}{lYY}
\toprule
\textbf{Definition} & \textbf{Positive evidence} & \textbf{Negative evidence} \\
\midrule
MoE FP8 & Kept the \(L=1\) epilogue and \(L=901\) route/pack specialization; mean latency improved from 0.299662 ms to 0.286342 ms with 19/19 passing. & The forced optional Triton backend was rejected: the forced-dispatch probe rose to 0.495863 ms mean latency despite correctness. \\
DSA Top-k & CuTe scorer tile size \(N=16\) reduced medium-band scorer latency from 6.783 us to 4.896 us and full-sweep mean from 0.007604 ms to 0.006893 ms. & Tile size \(N=32\) regressed because per-CTA work grew faster than the CTA-count reduction; a higher launch-bound setting had no measurable effect. \\
DSA Attention & Half-width tail skipping in the large split route improved the repeat mean from 0.011267 ms to 0.011175 ms with 23/23 passing. & Shared-memory carveout and launch-bound probes on a high-span workload were rejected (0.016640 ms and 0.015333 ms, respectively, versus the 0.013866 ms baseline). \\
GDN Decode & The batch-eight one-warp launch specialization improved all seven batch-eight workloads and reduced their mean from 0.004147 ms to 0.004050 ms. & Larger-batch changes did not generalize: batch 32 and 64 slightly regressed in the paired record, so those routes remained separate. \\
GDN Prefill & The Blackwell chunk path plus narrow short fallback reduced mean latency from 0.185183 ms to 0.051992 ms and improved 90/100 workloads. & The old broad 40-pair dispatch table was removed; only 15 tiny measured-regression shapes retained the fallback. \\
\bottomrule
\end{tabularx}
\end{table}

\Needspace{20\baselineskip}
\section{DSA Top-k Repeated-Validation Evidence}
\label{sec:appendix-dsa-topk-repeated-validation}

This appendix records the follow-up checks used to understand the DSA top-k indexer numerical issue observed after submission. The contest-side validation is treated as the authoritative scoring signal. The local retained result is reported only as development-time harness evidence, and the repeated-validation runs are used to explain the failure mode and the resulting harness policy.

The issue is a trial-budget problem rather than a latency-path mismatch. During normal agent search, running many trials for every candidate and every workload would make B200 iteration too slow and expensive, so the loop used representative gates and low-trial full sweeps to keep search moving. In the public FlashInfer-Bench configuration checked for this paper, the resolved default evaluation budget is \(n=3\) trials and the DSA top-k definition does not have a separate trial-count override~\citep{flashinferbench}. Each trial regenerates or reloads a distinct input/state sample before correctness is checked, so the default budget observes only three draws from the workload distribution. That policy is effective for throughput exploration, but it can produce apparent inconsistencies: a low-trial run may not sample a near-tie top-k boundary case, while another run or a high-trial replay can sample it and trigger the exact matched-ratio gate.

The detection effect can be modeled directly. For a fixed workload \(w\), let \(F_{w,t}\) denote the event that trial \(t\) violates the exact matched-ratio correctness gate, and let
\[
p_w = \Pr(F_{w,t}).
\]
Under an independent-trial approximation, a validation run with \(n\) trials detects at least one such failure with probability
\[
P_{\mathrm{detect}}(w,n)
=
1-(1-p_w)^n,
\qquad
P_{\mathrm{miss}}(w,n)
=
(1-p_w)^n .
\]
For a set of workloads \(S\), a low-trial full sweep misses all rare failures with probability approximately
\[
P_{\mathrm{miss}}(S,n)
=
\prod_{w\in S}(1-p_w)^n .
\]
This explains why a low-trial gate can pass while a targeted high-trial replay of the same suspicious workload later exposes a rare failure.

\begin{figure}[!htbp]
\centering
\includegraphics[width=0.92\textwidth]{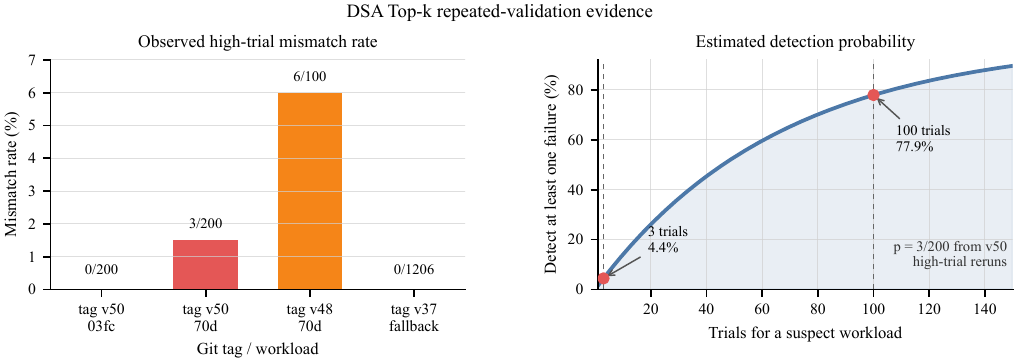}
\caption{\normalfont Observed high-trial mismatch rates and trial-budget detection probability for DSA top-k. The labels v50, v48, and v37 denote repository git tags, not iteration numbers; the probability curve uses the observed git tag v50 mismatch rate \(\hat p=3/200\).}
\label{fig:appendix-dsa-topk-repeated-validation}
\end{figure}

The top-k selector also has a natural stability condition. Let \(s_j\) be the reference score for page \(j\), \(T_k(s)\) the selected top-\(k\) set, and \(\hat{s}_j=s_j+\delta_j\) the implemented score after FP8 arithmetic, tiling, and reduction-order perturbations. Define the top-k boundary margin
\[
\gamma(s)
=
\min_{i\in T_k(s),\,j\notin T_k(s)}
|s_i-s_j|.
\]
If \(\max_j|\delta_j|\le \epsilon\) and \(\gamma(s)>2\epsilon\), then the selected set is stable: \(T_k(\hat{s})=T_k(s)\). Trial-sensitive failures concentrate in the boundary regime \(\gamma(s)\le 2\epsilon\), where small numerical perturbations can swap an in-set and out-of-set page while most trials remain exact. Figure~\ref{fig:appendix-dsa-topk-repeated-validation} visualizes the observed mismatch rates and the resulting detection-probability curve under the high-trial rerun evidence.

\begin{table}[H]
\centering
\caption{Concern workloads and repeated-validation outcomes for DSA top-k. The rows summarize archived B200 reruns; max error is reported as max absolute / max relative error.}
\label{tab:appendix-dsa-topk-repeated-validation}
\scriptsize
\setlength{\tabcolsep}{3pt}
\renewcommand{\arraystretch}{1.08}
\begin{tabularx}{\textwidth}{@{}p{0.09\textwidth}p{0.16\textwidth}p{0.07\textwidth}p{0.13\textwidth}p{0.12\textwidth}p{0.11\textwidth}p{0.11\textwidth}Y@{}}
\toprule
\textbf{UUID} & \textbf{Axes} & \textbf{\shortstack{Git\\tag}} & \textbf{Protocol} & \textbf{Status} & \textbf{\shortstack{Mismatches\\ / trials}} & \textbf{\shortstack{Max\\error}} & \textbf{Interpretation} \\
\midrule
\code{03fc111f} & \(B=4\), \(P_{\max}=45\) & v50 & official-style 3 & PASSED & 0/3 & 0 / 0 & Low-trial rerun did not reproduce the contest-side concern. \\
\code{03fc111f} & \(B=4\), \(P_{\max}=45\) & v50 & 100 trials, 2 reps & PASSED & 0/200 & 0.253 / 0.0079 & Nonzero differences appeared in sparse trials but remained within tolerance. \\
\code{70d53807} & \(B=12\), \(P_{\max}=82\) & v50 & official-style 3 & PASSED & 0/3 & 0 / 0 & Low-trial rerun missed the rare failure mode. \\
\code{70d53807} & \(B=12\), \(P_{\max}=82\) & v50 & 100 trials, 2 reps & INCORRECT & 3/200 & 6.033 / 0.1587 & High-trial replay reproduced sparse selector-boundary failures. \\
\code{70d53807} & \(B=12\), \(P_{\max}=82\) & v48 & 100 trials & INCORRECT & 6/100 & 7.280 / 0.1854 & The issue was workload-sensitive and not isolated to the final tag. \\
Target pair & two concern UUIDs & v37 & 3, 100, and 500 trials & PASSED & 0/1206 & 0 / 0 & Conservative fallback evidence: all repeated checks stayed exact. \\
\bottomrule
\end{tabularx}
\end{table}

\begin{table}[H]
\centering
\caption{DSA top-k latency consistency on contest-side timed rows. The two rows that failed correctness do not have contest-side latency rows, so this table compares only the reported timed rows.}
\label{tab:appendix-dsa-topk-latency-consistency}
\small
\setlength{\tabcolsep}{4pt}
\begin{tabular}{@{}lrrrrr@{}}
\toprule
\textbf{Set} & \textbf{Rows} & \textbf{\shortstack{Contest-side\\avg (ms)}} & \textbf{\shortstack{Local run 1\\avg (ms)}} & \textbf{\shortstack{Local run 2\\avg (ms)}} & \textbf{\shortstack{Contest vs.\\local mean}} \\
\midrule
All timed rows & 126 & 0.006785714 & 0.006813743 & 0.006765329 & -0.06\% \\
\(P_{\max}\) 1--32 & 69 & 0.002000000 & 0.001987188 & 0.002010386 & 0.06\% \\
\(P_{\max}\) 33--36 & 18 & 0.011111111 & 0.011295343 & 0.010980130 & -0.24\% \\
\(P_{\max}\) 37--45 & 20 & 0.011400000 & 0.011553300 & 0.011395958 & -0.65\% \\
\(P_{\max}\) 82--91 & 19 & 0.015210526 & 0.015107026 & 0.015165965 & 0.49\% \\
\bottomrule
\end{tabular}
\end{table}

These checks changed the harness policy rather than the paper's scoring interpretation. Low-trial validation remains appropriate inside the agent search loop because it preserves iteration speed, but suspicious DSA top-k regimes now require targeted high-trial replay before a tag is treated as the most conservative release artifact.

\Needspace{18\baselineskip}
\section{Operator Background and Mathematical Definitions}
\label{sec:appendix-operator-math}

The contest page defines the three tracks as FP8 Fused MoE, DeepSeek Sparse Attention from DeepSeek-V3.2, and Gated Delta Net used in Qwen3-Next~\citep{contest}. This appendix records the operator-level mathematical background used to interpret the benchmark definitions. The formulas are not new model contributions; they explain why the kernels stress routing, top-k selection, sparse memory access, and recurrent state updates.

\subsection{Fused MoE FP8}

The MoE task is naturally described by top-k routed expert execution. In a Transformer block where a dense FFN is replaced by an MoE layer, DeepSeekMoE writes the token update as~\citep{deepseekmoe}
\[
h_t^\ell
=
u_t^\ell
+
\sum_{i=1}^{N}
g_{i,t}\,\mathrm{FFN}_i(u_t^\ell),
\qquad
g_{i,t}
=
\begin{cases}
s_{i,t}, & s_{i,t}\in \mathrm{TopK}(\{s_{j,t}\}_{j=1}^{N}, K),\\
0, & \text{otherwise},
\end{cases}
\]
\[
s_{i,t}
=
\mathrm{softmax}_i\!\left((u_t^\ell)^\top e_i^\ell\right).
\]
Here $u_t^\ell$ is the token hidden state after attention, $e_i^\ell$ is the router centroid for expert $i$, and only $K$ expert FFNs are active per token. The contest definition specializes this pattern to a fixed top-8 routed FP8 MoE with block-scale quantization. At the kernel level, the mathematical sparsity becomes a systems problem: gather routed tokens, pack expert-major work, run two expert GEMMs with FP8 block scales, apply the activation/epilogue, and scatter weighted expert outputs back to token order.

For block-scaled FP8 inputs and weights, each tile is represented as a low-precision tensor and a scale. Abstractly, a GEMM tile computes
\[
C_{m,n}
\approx
\sum_k
\left(s^x_{b(m,k)} q^x_{m,k}\right)
\left(s^w_{b(k,n)} q^w_{k,n}\right),
\]
where $q^x,q^w$ are FP8 values and $s^x,s^w$ are per-block scales. This is why the retained MoE implementation emphasized FP8 scale handling, expert-major packing, and epilogue fusion rather than only raw matrix multiplication throughput.

\subsection{DeepSeek Sparse Attention}

DeepSeek-V3.2 introduces DSA to reduce long-context attention cost by learning a lightweight indexer and applying attention only to selected key-value entries~\citep{deepseekv32}. For query token $h_t$ and preceding token $h_s$, the lightning indexer computes
\[
I_{t,s}
=
\sum_{j=1}^{H_I}
w^I_{t,j}\,
\mathrm{ReLU}\!\left((q^I_{t,j})^\top k^I_s\right),
\]
where $H_I$ is the number of indexer heads, $q^I_{t,j}$ and $w^I_{t,j}$ are query-derived indexer features, and $k^I_s$ is a key-derived indexer feature. The selected context set is
\[
S_t
=
\{s \mid I_{t,s}\in \mathrm{TopK}(I_{t,:}, k)\}.
\]
The sparse attention output is then computed only over the selected entries:
\[
u_t
=
\mathrm{Attn}\!\left(h_t,\{c_s \mid s\in S_t\}\right).
\]

In the contest, this decomposition appears as two definitions. The DSA top-k indexer kernel computes index scores and selected indices across \code{batch\_size}, \code{max\_num\_pages}, and fixed \code{num\_pages}. The DSA sparse attention kernel consumes those sparse indices and performs attention over the selected KV pages. In kernel notation, the attention route can be summarized as
\[
a_{t,s}
=
\mathrm{softmax}_{s\in S_t}
\left(
\lambda
\left[
q^{\mathrm{nope}}_t{}^\top c^{K}_s
+
q^{\mathrm{pe}}_t{}^\top k^{\mathrm{pe}}_s
\right]
\right),
\qquad
o_t
=
\sum_{s\in S_t} a_{t,s} c^{V}_s,
\]
where $\lambda$ is the softmax scale, $q^{\mathrm{nope}}$ and $q^{\mathrm{pe}}$ are the non-positional and positional query components, and $c^K,c^V,k^{\mathrm{pe}}$ denote the corresponding cached latent/value and positional-key components. The exact tensor layout is fixed by the contest reference implementation; the optimization challenge is to preserve the selected-set semantics while avoiding wasted work on padded pages and short-token regimes.

\subsection{Gated Delta Net}

Gated DeltaNet belongs to the linear-recurrent attention family. A simple linear attention recurrence maintains a matrix state $S_t$ and emits
\[
S_t = S_{t-1} + v_t k_t^\top,
\qquad
o_t = S_t q_t .
\]
DeltaNet replaces the additive write with a selective update that removes the old value associated with key $k_t$ before writing the new value:
\[
S_t
=
S_{t-1}(I-\beta_t k_t k_t^\top)
+
\beta_t v_t k_t^\top .
\]
Gated DeltaNet adds a data-dependent decay gate $\alpha_t\in(0,1)$, yielding the gated delta rule~\citep{gdn}:
\[
S_t
=
S_{t-1}\left(\alpha_t(I-\beta_t k_t k_t^\top)\right)
+
\beta_t v_t k_t^\top,
\qquad
o_t = S_t q_t .
\]

The two contest definitions correspond to two execution regimes of this recurrence. Decode receives a prior state and computes a small number of new recurrent updates, so launch overhead, batch-size dispatch, and state layout dominate. Prefill evaluates a whole sequence and returns both outputs and the final state, so the key issue is parallelizing the recurrence through chunked matrix operations while applying the log-domain gate consistently, e.g. $\alpha_t=\exp(a_t)$ when the reference exposes a natural-log gate. This explains why the retained GDN prefill solution benefited from a Blackwell chunked path, while GDN decode used batch-size-specific dispatch.

\Needspace{18\baselineskip}
\section{Agent-Assisted Profiling Evidence}
\label{sec:appendix-profile-evidence}

This appendix summarizes the largest-shape profiling evidence requested by the review committee. Each profile uses the final Agent-Assisted retained implementation for the selected workload. The table keeps only the Torch Profiler and NCU facts that affected optimization decisions: dominant CUDA time, kernel-level throughput, occupancy or wave count, and the resulting implication. The corresponding artifact archive contains the raw Torch Profiler tables, NCU reports, selected-workload metadata, evaluator outputs, and run logs; those raw logs are summarized here rather than inlined.

\begin{table}[!htbp]
\centering
\caption{Condensed largest-shape profiling evidence for the retained Agent-Assisted kernels. The table summarizes raw Torch Profiler and NCU reports from the artifact archive.}
\label{tab:appendix-profile-summary}
\scriptsize
\setlength{\tabcolsep}{3pt}
\renewcommand{\arraystretch}{1.08}
\begin{tabularx}{\textwidth}{>{\raggedright\arraybackslash}p{0.13\textwidth} >{\raggedright\arraybackslash}p{0.18\textwidth} r Y Y}
\toprule
\textbf{Operator} & \textbf{Regime} & \textbf{\shortstack{Retained workload\\latency (ms)}} & \textbf{Dominant profiler signal} & \textbf{NCU-backed decision} \\
\midrule
MoE FP8 & long seq (L=14107) & 1.189900 & gemm1 50.8\%, gemm2 30.8\% of CUDA time; auxiliary routing/pack/epilogue kernels make up the remaining 18.4\% & one wave/SM; reg/shared-memory-limited GEMM occupancy; tune expert GEMMs together with pack/reduce traffic \\
DSA Top-k Indexer & high page-count (B=30, Pmax=91) & 0.017546 & selector 55.3\% and CuTe scorer 44.8\% of CUDA time & scorer 0.67 waves/SM; selector one-block 0.01 waves/SM; move next effort to selector parallelism or launch/fusion reduction \\
DSA Attention & T=8 sparse attention & 0.015115 & split 71.9\% and merge 28.1\% of CUDA time & split/merge are low-wave sparse launches, not bandwidth-saturated; keep tail skipping; focus next on split/merge scheduling \\
GDN Decode & batch=64 decode & 0.012570 & one CuTe pretranspose decode kernel accounts for 100\% of CUDA time & high occupancy but only 1.73 waves/SM; moderate SM/memory throughput; keep batch-specialized dispatch instead of replacing the route \\
GDN Prefill & 8192-token prefill & 0.134540 & Blackwell chunk kernel 98.5\% of CUDA time; fused gate helper 1.5\% & chunk kernel is reg/shared-memory limited with spills and coalescing headroom; keep Blackwell chunk default; future work targets resources/coalescing \\
\bottomrule
\end{tabularx}
\end{table}

\subsection{Detailed Per-Operator Analysis}

The NCU kernel durations below describe individual profiled kernel passes; they are used as bottleneck evidence and are not summed into end-to-end latency.

\medskip
\noindent\textbf{MoE FP8.} \textbf{Workload:}\ Selected stress workload (seq\_len=14107) covers the long seq (L=14107) regime. The retained latency is 1.189900 ms; the short profiling-run latency is 1.172120 ms.

\noindent\textbf{Torch Profiler:}\ Torch Profiler reports that gemm1\_kernel takes 627.519 us, or 50.78\% of CUDA time, and gemm2\_kernel takes 380.639 us, or 30.80\%. The two expert GEMMs therefore account for 81.6\% of CUDA time. The remaining time is still meaningful: reduce\_add contributes 6.53\%, fusedGating 4.21\%, FP8 permute/pack 3.79\%, and activation quantization 2.78\%.

\noindent\textbf{NCU:}\ NCU shows that the two GEMM kernels launch one block per SM with 384 threads per block. Both use 168 registers per thread and roughly 216 KB of dynamic shared memory per block, which limits achieved occupancy to about 18.7\%. GEMM1 reaches 50.2\% SM throughput and 41.1\% memory throughput, while GEMM2 reaches 37.8\% SM throughput and 32.8\% memory throughput. This is a mixed tensor-memory and movement bottleneck rather than a launch-only case.

\noindent\textbf{Optimization implication:}\ The profile supports keeping the long-sequence route and optimizing the expert-major GEMM path together with FP8 packing, scale movement, and reduction/epilogue work. Treating the epilogue as an isolated microkernel would miss the dominant cost structure.

\medskip
\noindent\textbf{DSA Top-k Indexer.} \textbf{Workload:}\ Selected stress workload (batch\_size=30, max\_num\_pages=91, num\_pages=11923) covers the high page-count (B=30, Pmax=91) regime. The retained latency is 0.017546 ms; the short profiling-run latency is 0.015168 ms.

\noindent\textbf{Torch Profiler:}\ Torch Profiler shows only two meaningful CUDA kernels on the largest shape: the filtered single-row top-k selector takes 8.928 us, or 55.25\% of CUDA time, and the CuTe scorer takes 7.232 us, or 44.75\%. The final largest-shape profile is therefore no longer scorer-only.

\noindent\textbf{NCU:}\ NCU reports that the CuTe scorer launches 394 blocks, reaches 18.9\% SM throughput and 31.3\% memory throughput, and executes only 0.67 waves per SM. The selector is more extreme: it is a one-block launch with 0.01 waves per SM and near-zero device-wide compute and memory utilization. This explains why the selector can consume more CUDA time despite not saturating the device.

\noindent\textbf{Optimization implication:}\ The earlier scorer tile optimization remains justified, but the next largest-shape bottleneck is selector underfill and launch overhead. Promising follow-up directions are selector parallelization, reducing launch count, or fusing scorer and selector work where correctness permits.

\medskip
\noindent\textbf{DSA Attention.} \textbf{Workload:}\ Selected stress workload (num\_tokens=8, num\_pages=8462, max\_pages=32) covers the T=8 sparse attention regime. The retained latency is 0.015115 ms; the short profiling-run latency is 0.015552 ms.

\noindent\textbf{Torch Profiler:}\ Torch Profiler attributes 8.416 us, or 71.86\% of CUDA time, to attn\_split\_kernel and 3.296 us, or 28.14\%, to attn\_merge\_kernel. The largest sparse-attention case is therefore a short two-kernel pipeline rather than a single dense attention kernel.

\noindent\textbf{NCU:}\ NCU shows the split kernel uses 126 registers per thread and 181 KB of dynamic shared memory per block, with 12.5\% theoretical occupancy, 12.0\% achieved occupancy, and 0.86 waves per SM. The merge kernel is even smaller, with 0.09 waves per SM and about 3.1\% achieved occupancy. Both kernels show low device-wide throughput; the profile points to sparse-grid underfill and resource pressure rather than DRAM saturation.

\noindent\textbf{Optimization implication:}\ The retained half-width tail skip is supported because it removes real padded work in the large sparse route. The remaining headroom is more likely in split/merge scheduling, reducing merge overhead, or carefully fusing pipeline stages than in generic bandwidth tuning.

\medskip
\noindent\textbf{GDN Decode.} \textbf{Workload:}\ Selected stress workload (batch\_size=64) covers the batch=64 decode regime. The retained latency is 0.012570 ms; the short profiling-run latency is 0.012352 ms.

\noindent\textbf{Torch Profiler:}\ Torch Profiler reports one CUDA kernel carrying the device work on the largest decode shape: the CuTe pretranspose decode kernel takes 9.952 us and accounts for 100\% of CUDA time.

\noindent\textbf{NCU:}\ NCU measures the profiled decode kernel at about 12.8 us, with 44.0\% SM throughput, 52.1\% memory throughput, and 34.6\% DRAM throughput. Achieved occupancy is high at roughly 82.9\%, but the launch executes only 1.73 waves per SM and NCU flags a partial-wave tail effect. The limiter is therefore latency and wave shape, not a saturated tensor-core or DRAM kernel.

\noindent\textbf{Optimization implication:}\ The evidence supports the final dispatch design: batch-local routes are useful, but the largest batch already uses a strong CuTe path. A wholesale replacement is unlikely to dominate the full workload distribution unless it also improves state access and tail-wave behavior.

\medskip
\noindent\textbf{GDN Prefill.} \textbf{Workload:}\ Selected stress workload (total\_seq\_len=8192, num\_seqs=57, len\_cu\_seqlens=58) covers the 8192-token prefill regime. The retained latency is 0.134540 ms; the short profiling-run latency is 0.141919 ms.

\noindent\textbf{Torch Profiler:}\ Torch Profiler shows that the retained Blackwell chunk kernel takes 130.111 us, or 98.47\% of CUDA time. The fused gate helper takes only 2.016 us, or 1.53\%, so the helper is not the largest-shape bottleneck.

\noindent\textbf{NCU:}\ NCU reports the chunk kernel at about 133.3 us, with 8.5\% SM throughput and 24.0\% memory throughput. The kernel uses 168 registers per thread and about 100 KB of dynamic shared memory per block; both registers and shared memory limit occupancy. Theoretical occupancy is 18.75\% and achieved occupancy is about 16.2\%. NCU also reports 976 local spilling requests and 3,382,832 excessive global sectors, about 50\% of total sectors.

\noindent\textbf{Optimization implication:}\ This supports the current Blackwell chunk path as the right default for non-tiny prefill shapes, because the dominant work is the chunked recurrent kernel rather than gate preprocessing. Future tuning should focus on register/shared-memory pressure and global-memory coalescing before adding more dispatch cases.

\Needspace{18\baselineskip}
\section{Per-Workload Statistics and Shape Tables}

\label{sec:appendix-stats}

The statistics in this appendix are computed from retained per-workload artifacts joined with contest workload axes. In Table~\ref{tab:appendix-distribution}, \emph{Workload} denotes the number of retained evaluated workloads for each definition. The 95th percentile is computed across retained per-workload latency values, not trial-level variance. For DSA top-k, the promoted full-sweep summary records 0.006893 ms, while the available per-workload artifact used in this appendix has mean 0.007649 ms.

\begin{table}[H]
\centering
\caption{Retained per-workload latency statistics from available artifacts. Workloads is the retained workload count; latency values are milliseconds; artifact mean is computed from the available per-workload artifact; PyTorch mean is the reference mean from the same evaluator output.}
\label{tab:appendix-distribution}
\setlength{\tabcolsep}{4pt}
\begin{tabular}{@{}lrrrrrrr@{}}
\toprule
\textbf{Definition} & \textbf{Workloads} & \textbf{\shortstack{Artifact\\mean}} & \textbf{Median} & \textbf{P95} & \textbf{Min} & \textbf{Max} & \textbf{\shortstack{PyTorch\\mean}} \\
\midrule
MoE FP8 & 19 & 0.28634 & 0.22745 & 0.89852 & 0.06693 & 1.210 & 18.262 \\
DSA Top-k Indexer & 128 & 0.007649 & 0.002187 & 0.01756 & 0.001696 & 0.01838 & 3.406 \\
DSA Sparse Attention & 23 & 0.01113 & 0.01302 & 0.01526 & 0.005248 & 0.01530 & 2.427 \\
GDN Decode & 54 & 0.006201 & 0.004910 & 0.01257 & 0.002420 & 0.01276 & 49.420 \\
GDN Prefill & 100 & 0.05199 & 0.02105 & 0.15775 & 0.005470 & 0.23132 & 1095.884 \\
\bottomrule
\end{tabular}
\end{table}

\begin{figure}[H]

\centering

\includegraphics[width=0.88\textwidth]{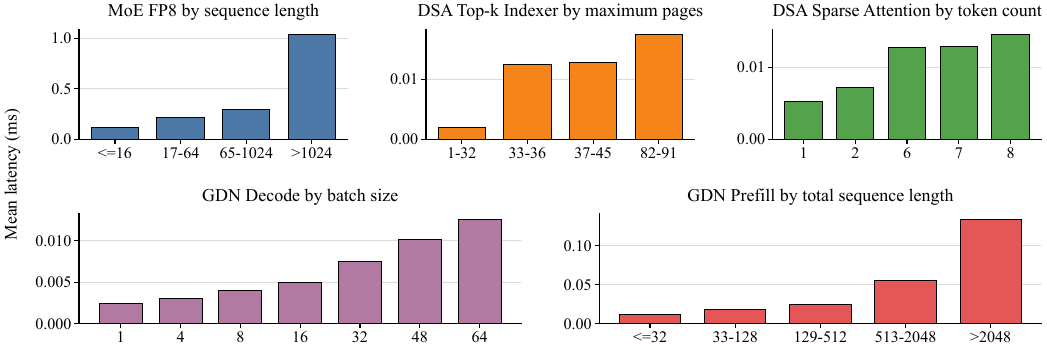}

\caption{Mean retained latency by primary workload axis; high-cardinality axes are bucketed into workload-relevant ranges.}

\label{fig:appendix-axis-means}

\end{figure}

\noindent\textbf{Primary-axis mean view.}
Figure~\ref{fig:appendix-axis-means} groups each definition by the shape axis that most directly changes its execution regime: sequence length for MoE and GDN prefill, maximum page count for DSA top-k, token count for DSA sparse attention, and batch size for GDN decode. Each bar reports mean retained latency for that bucket; the UUID-level tables below preserve the exact shapes and measurements.

\section{Per-Shape Retained Results}

\label{sec:appendix-shape-results}

\setlength{\LTcapwidth}{\textwidth}

\setlength{\LTleft}{\fill}

\setlength{\LTright}{\fill}

Each table below corresponds to one contest definition. Each row is one retained workload joined with shape axes from the contest workload file; repeated shapes with different input data remain separate. The tables use full workload UUIDs and report retained latency, PyTorch reference latency, and speedup over PyTorch. Per-row PyTorch-reference speedups are diagnostic and are not the FlashInfer-baseline speedups reported in the main results.

\Needspace{12\baselineskip}

\subsection{MoE FP8}

{\scriptsize
\setlength{\tabcolsep}{2pt}
\setlength{\abovecaptionskip}{2pt}
\setlength{\belowcaptionskip}{3pt}
\setlength{\LTpre}{2pt}
\setlength{\LTpost}{3pt}
\renewcommand{\arraystretch}{0.88}
\begin{longtable}{@{}>{\raggedright\arraybackslash}p{0.42\textwidth}>{\raggedleft\arraybackslash}p{0.13\textwidth}>{\raggedleft\arraybackslash}p{0.12\textwidth}>{\raggedleft\arraybackslash}p{0.12\textwidth}>{\raggedleft\arraybackslash}p{0.12\textwidth}@{}}
\caption{Per-workload retained results for MoE FP8. Latency and PyTorch reference are milliseconds; speedup is relative to PyTorch.}
\label{tab:appendix-shapes-moe-fp8}\\
\toprule
\textbf{Workload UUID} & \textbf{\shortstack{Sequence\\length}} & \textbf{Latency} & \textbf{\shortstack{PyTorch\\reference}} & \textbf{\shortstack{Speedup vs.\\PyTorch}} \\
\midrule
\endfirsthead
\multicolumn{5}{@{}l}{\textit{Table~\thetable{} continued from previous page}} \\
\toprule
\textbf{Workload UUID} & \textbf{\shortstack{Sequence\\length}} & \textbf{Latency} & \textbf{\shortstack{PyTorch\\reference}} & \textbf{\shortstack{Speedup vs.\\PyTorch}} \\
\midrule
\endhead
e05c6c03-5603-4a1c-b34c-dcce0ecaeea4 & 1 & 0.06693 & 11.733 & 175.3$\times$ \\
b8f4f012-a32e-4356-b4e1-7665b3d598af & 7 & 0.10348 & 12.546 & 121.2$\times$ \\
8cba5890-4288-448a-93b8-42c14c6b9420 & 14 & 0.15226 & 13.729 & 90.2$\times$ \\
2e69caee-ae5c-473b-aa99-5dc6659829d4 & 15 & 0.08786 & 12.222 & 139.1$\times$ \\
a7c2bcfd-a2f4-479e-8d32-200115df89cf & 16 & 0.15449 & 13.976 & 90.5$\times$ \\
6230e838-67ca-41dd-a9d6-6f36b7676c6b & 32 & 0.22283 & 15.839 & 71.1$\times$ \\
f7d6ac7c-24ec-47e4-aefc-4863a5e3e1d9 & 52 & 0.18008 & 14.831 & 82.4$\times$ \\
fc378037-e8fa-4305-b00f-4af47933fd53 & 53 & 0.22745 & 16.407 & 72.1$\times$ \\
76010cb4-f73c-4145-8365-8642a2ce99de & 54 & 0.22793 & 16.175 & 71.0$\times$ \\
81955b1e-086f-49c1-9f40-a18a5aaf509d & 55 & 0.22933 & 16.411 & 71.6$\times$ \\
4822167c-dae5-4bb1-bb53-e4adb256245b & 56 & 0.24820 & 16.871 & 68.0$\times$ \\
74d7ff04-0365-4cf8-a824-ce61b7131dea & 57 & 0.23279 & 16.646 & 71.5$\times$ \\
e626d3e6-6c29-4fd8-bb7c-5b09eec61702 & 58 & 0.23868 & 16.933 & 70.9$\times$ \\
eedc63b2-c03b-4cf2-8d16-1f46566e3af7 & 59 & 0.21390 & 15.279 & 71.4$\times$ \\
5eadab1e-a0e3-4966-b0fd-1115cd77497c & 62 & 0.18530 & 15.229 & 82.2$\times$ \\
8f1ff9f1-6747-41d1-a1d8-2868cdacf893 & 80 & 0.27073 & 17.535 & 64.8$\times$ \\
1a4c6ba1-3cd2-4d7d-b716-84f2d52b69fc & 901 & 0.32424 & 21.043 & 64.9$\times$ \\
58a34f27-7995-4155-8b46-f60a7225e20e & 11948 & 0.86389 & 36.933 & 42.8$\times$ \\
5e8dc11c-f2a9-42d5-8dce-9419cbf34d5d & 14107 & 1.210 & 46.639 & 38.5$\times$ \\
\bottomrule
\end{longtable}
}

\Needspace{12\baselineskip}

\subsection{DSA Top-k Indexer}

{\scriptsize
\setlength{\tabcolsep}{2pt}
\setlength{\abovecaptionskip}{2pt}
\setlength{\belowcaptionskip}{3pt}
\setlength{\LTpre}{2pt}
\setlength{\LTpost}{3pt}
\renewcommand{\arraystretch}{0.88}
\begin{longtable}{@{}>{\raggedright\arraybackslash}p{0.36\textwidth}>{\raggedleft\arraybackslash}p{0.105\textwidth}>{\raggedleft\arraybackslash}p{0.105\textwidth}>{\raggedleft\arraybackslash}p{0.115\textwidth}>{\raggedleft\arraybackslash}p{0.115\textwidth}>{\raggedleft\arraybackslash}p{0.115\textwidth}@{}}
\caption{Per-workload retained results for DSA Top-k Indexer. Latency and PyTorch reference are milliseconds; speedup is relative to PyTorch. Fixed: page count = 11923.}
\label{tab:appendix-shapes-dsa-top-k-indexer}\\
\toprule
\textbf{Workload UUID} & \textbf{\shortstack{Batch\\size}} & \textbf{\shortstack{Maximum\\pages}} & \textbf{Latency} & \textbf{\shortstack{PyTorch\\reference}} & \textbf{\shortstack{Speedup vs.\\PyTorch}} \\
\midrule
\endfirsthead
\multicolumn{6}{@{}l}{\textit{Table~\thetable{} continued from previous page}} \\
\toprule
\textbf{Workload UUID} & \textbf{\shortstack{Batch\\size}} & \textbf{\shortstack{Maximum\\pages}} & \textbf{Latency} & \textbf{\shortstack{PyTorch\\reference}} & \textbf{\shortstack{Speedup vs.\\PyTorch}} \\
\midrule
\endhead
30cecff1-7ea4-474b-90fc-7f4a87206d8e & 1 & 1 & 0.001989 & 0.93197 & 468.6$\times$ \\
cd594d26-4ac9-4b15-9305-f4c65c789c16 & 1 & 2 & 0.002037 & 0.93493 & 458.9$\times$ \\
44ddaa65-5d63-41d6-8849-7975b5f6efd5 & 1 & 3 & 0.002053 & 1.114 & 542.4$\times$ \\
4667f9ad-92f3-4503-9eea-547a30e708dc & 2 & 1 & 0.001936 & 1.061 & 548.1$\times$ \\
b2098949-a0e4-4de5-9797-3d338dbc54b0 & 2 & 2 & 0.001968 & 1.296 & 658.8$\times$ \\
545f8a85-3fe9-4604-898b-8b21b2cfdf11 & 2 & 3 & 0.001877 & 1.287 & 685.8$\times$ \\
0ebafac4-a597-4245-9ea9-14089ac4b8b3 & 2 & 4 & 0.001952 & 1.072 & 549.2$\times$ \\
02fa7f90-88b6-4fa5-ba21-29b803c23309 & 2 & 5 & 0.001978 & 1.305 & 659.8$\times$ \\
8f2fde6c-585d-48fd-86cb-3a004c2b7884 & 2 & 6 & 0.001845 & 1.604 & 869.2$\times$ \\
e49574dd-5ee4-4e19-be85-fe41abae9a95 & 2 & 7 & 0.001904 & 1.094 & 574.7$\times$ \\
82a8a885-ad69-4b9e-a1a9-dd9c0715db91 & 2 & 8 & 0.001792 & 1.297 & 723.7$\times$ \\
83cb81c5-7304-449a-bead-8ce4a444e2df & 3 & 2 & 0.001760 & 1.217 & 691.6$\times$ \\
46f236c0-3aef-4628-ae2a-3b313986b9eb & 4 & 1 & 0.001771 & 1.345 & 759.4$\times$ \\
d54c1568-87bd-4edf-a329-f0ee35091e45 & 4 & 2 & 0.002021 & 1.340 & 663.2$\times$ \\
55be3dc3-b010-40b9-9591-c5349849f4e9 & 4 & 3 & 0.001733 & 1.343 & 774.8$\times$ \\
899c2d2f-540c-432f-9944-e29d4cc6fe87 & 4 & 4 & 0.001754 & 1.322 & 753.6$\times$ \\
9754a4e7-3db5-4975-9ef4-3f3b6c61dc97 & 4 & 16 & 0.001754 & 1.760 & 1003.1$\times$ \\
4279d75e-b93c-4198-9016-4d1d21e17bf2 & 4 & 17 & 0.002080 & 1.382 & 664.6$\times$ \\
05775386-52aa-4abb-982d-0aee72dd2ff6 & 4 & 18 & 0.002080 & 1.418 & 681.7$\times$ \\
28a9fa48-97c3-4d60-b7c2-ffeb230c9f1a & 4 & 19 & 0.002000 & 1.353 & 676.7$\times$ \\
17ced9b8-339a-4725-ad28-ce52a0fead0f & 4 & 30 & 0.001845 & 1.930 & 1046.1$\times$ \\
101a39ac-6465-45d8-88c5-b58dc3bc6398 & 4 & 31 & 0.002069 & 1.335 & 645.1$\times$ \\
ef0d0deb-ac85-4d72-90ce-a811f2774c10 & 4 & 32 & 0.002208 & 1.653 & 748.6$\times$ \\
d8a73470-cfa1-49ad-aba6-d9b99c0c5b2d & 4 & 34 & 0.01214 & 1.873 & 154.3$\times$ \\
cd3434ac-bdda-4c73-98ff-c746b236b856 & 4 & 35 & 0.01182 & 1.341 & 113.4$\times$ \\
4c7705ad-0954-4628-a978-a51a895aa008 & 4 & 36 & 0.01200 & 1.351 & 112.6$\times$ \\
16feeab1-509f-49f2-b794-4ed42b843984 & 4 & 43 & 0.01291 & 1.345 & 104.2$\times$ \\
ef12ac76-832a-439b-9c87-ce33f311e125 & 4 & 44 & 0.01306 & 1.741 & 133.4$\times$ \\
03fc111f-0f63-4ca1-9765-cbaca21ada2b & 4 & 45 & 0.01339 & 1.880 & 140.4$\times$ \\
9410ad1e-e4f6-4911-bc96-0931dcb00838 & 6 & 1 & 0.001776 & 1.651 & 929.7$\times$ \\
9a2bb7f8-b1f9-43d6-91c4-cafac5ec581e & 6 & 26 & 0.001856 & 2.517 & 1355.9$\times$ \\
09bb020f-60d4-4393-a3df-765fd51011d4 & 6 & 28 & 0.002053 & 1.601 & 780.0$\times$ \\
10b4eebe-e458-410a-8b99-2784429c43b2 & 6 & 32 & 0.002165 & 2.402 & 1109.5$\times$ \\
df80c00b-9bc2-4fac-82aa-101301d4c2dd & 7 & 18 & 0.001813 & 2.416 & 1332.7$\times$ \\
e515e20a-6dd7-44fe-9094-56be36424806 & 7 & 32 & 0.001925 & 2.404 & 1248.7$\times$ \\
99920dc5-e1af-4083-a358-66ea4f897e05 & 7 & 34 & 0.01247 & 1.751 & 140.4$\times$ \\
e64a4ebc-1b1b-4ec3-8c67-a9a87533946c & 8 & 1 & 0.001733 & 1.851 & 1068.3$\times$ \\
13dad24c-6482-436a-bb10-b54c3280a0a6 & 8 & 2 & 0.001744 & 1.828 & 1048.3$\times$ \\
8f1a5846-5499-40c7-9904-6887fda49a80 & 8 & 3 & 0.001696 & 2.573 & 1517.2$\times$ \\
3240d5fa-9321-4bc1-b4bf-225416da1e2c & 8 & 8 & 0.001765 & 2.768 & 1568.0$\times$ \\
c729310b-1f3a-4245-ac81-ac366d57b0e3 & 8 & 9 & 0.002016 & 1.919 & 951.8$\times$ \\
82bd3e70-4d7d-44a0-b97e-b5db535fca9e & 8 & 16 & 0.001867 & 2.812 & 1506.3$\times$ \\
f897f64e-68b7-4f23-a2c5-bba516828f43 & 8 & 17 & 0.001888 & 1.934 & 1024.2$\times$ \\
6caf09cf-7a2e-47e0-96fa-67807366fd37 & 8 & 18 & 0.001888 & 2.682 & 1420.6$\times$ \\
9c313fc4-4453-4dc8-b4c6-ccfde2148490 & 8 & 19 & 0.001941 & 2.733 & 1408.0$\times$ \\
7752dda1-9d4d-4383-813b-c125ad3d038a & 8 & 26 & 0.002016 & 2.712 & 1345.4$\times$ \\
d04ea89f-d4c1-4fb5-8313-fb17cac68f75 & 8 & 27 & 0.001888 & 2.726 & 1443.6$\times$ \\
06ec358c-604b-4ca6-acd5-9c1953750fb3 & 8 & 28 & 0.002027 & 1.888 & 931.7$\times$ \\
a4cdaee6-2c3e-46fa-a60a-fef4ccf4c30b & 8 & 30 & 0.002027 & 2.455 & 1211.5$\times$ \\
1152c61f-09db-4fc6-b22a-698247cd8ce6 & 8 & 31 & 0.001920 & 2.622 & 1365.8$\times$ \\
67216408-b49b-4e43-bf07-4136968b3b82 & 8 & 32 & 0.002080 & 1.891 & 909.2$\times$ \\
b017f77a-7e57-4ab6-aa43-b82eabc3812f & 8 & 34 & 0.01220 & 2.415 & 198.0$\times$ \\
8638fe06-e733-482f-99b8-946abcdfad11 & 8 & 35 & 0.01237 & 3.472 & 280.6$\times$ \\
8f3fe9ff-ce2a-4262-96e8-b2190f763fbb & 8 & 36 & 0.01267 & 2.709 & 213.8$\times$ \\
2774963f-026a-4284-9189-228676fc3f70 & 8 & 38 & 0.01242 & 3.020 & 243.1$\times$ \\
3e91afa0-0ef1-4e98-8ed5-6517c131c9f0 & 8 & 43 & 0.01321 & 1.942 & 147.1$\times$ \\
77279062-1275-41d8-9184-210994a8b4d7 & 8 & 44 & 0.01313 & 1.907 & 145.3$\times$ \\
81a953ea-f29f-4707-8456-67a3624a549d & 8 & 45 & 0.01312 & 1.870 & 142.5$\times$ \\
5f7e6f22-1dae-456a-80d1-a78bcc5e071e & 11 & 32 & 0.002229 & 3.390 & 1520.6$\times$ \\
70d53807-f4ba-4c16-a0be-8a1f9137e8e6 & 12 & 82 & 0.01704 & 3.637 & 213.5$\times$ \\
d0c00dd5-819f-490f-9f88-ead63698e9ee & 14 & 1 & 0.001813 & 2.621 & 1445.6$\times$ \\
7f03b670-842f-4394-b4be-07c2f68b1c6d & 14 & 32 & 0.002112 & 3.897 & 1845.3$\times$ \\
f59fd3e2-a6b0-4ce7-bddd-fa47b35b9b90 & 14 & 35 & 0.01331 & 3.887 & 292.0$\times$ \\
7f1cd9c2-1097-43de-8fc8-1fa63763cf9b & 14 & 40 & 0.01284 & 4.056 & 315.8$\times$ \\
e1a185dc-a520-490b-8f93-e6fff4583e1f & 14 & 84 & 0.01710 & 3.980 & 232.7$\times$ \\
f7f61b05-f6fc-4269-8f38-05ed599d51c3 & 14 & 91 & 0.01779 & 3.909 & 219.7$\times$ \\
752c2ee5-9450-4ca4-815a-fa3e00e07991 & 15 & 1 & 0.001877 & 2.642 & 1407.2$\times$ \\
1ece7fb3-be8d-42d3-923d-f5a85846ddee & 15 & 18 & 0.001941 & 2.914 & 1501.3$\times$ \\
97a6d5c2-ed84-4738-8073-f07cda4c5bd3 & 15 & 19 & 0.002016 & 4.621 & 2292.1$\times$ \\
67c09e9c-3f31-485f-9aef-983979a90038 & 15 & 30 & 0.001995 & 4.220 & 2115.7$\times$ \\
e977c163-09a3-42aa-9d38-e629b67ee46a & 15 & 31 & 0.002070 & 2.818 & 1361.6$\times$ \\
bb22d09a-7fe9-46a1-a3af-90e636c3b966 & 15 & 32 & 0.002218 & 4.300 & 1938.2$\times$ \\
9a95a10e-bd5a-406a-a1d1-891c941343ce & 15 & 34 & 0.01254 & 3.836 & 305.9$\times$ \\
08a752fc-53b8-4335-b08f-ddb2c71eae65 & 15 & 35 & 0.01220 & 4.861 & 398.5$\times$ \\
175849a8-a360-4f8f-ac2b-88d278852d45 & 15 & 36 & 0.01274 & 2.863 & 224.7$\times$ \\
f1fc35d4-3052-49b2-ad6e-1f715a61a4cb & 15 & 38 & 0.01284 & 4.207 & 327.6$\times$ \\
f457feb2-6bf5-4be0-952d-580bf0b5f8b9 & 15 & 39 & 0.01301 & 4.397 & 338.0$\times$ \\
60605091-410d-47be-8764-9a516a256aa8 & 15 & 40 & 0.01321 & 2.846 & 215.4$\times$ \\
ee603b53-f9d6-4210-a83a-caf6219a8a8b & 15 & 43 & 0.01295 & 2.874 & 221.9$\times$ \\
e26d02ef-5606-4bbf-a397-d9a75d41312b & 15 & 44 & 0.01310 & 4.078 & 311.4$\times$ \\
b83c4150-dbc3-4d05-a2bf-92b7202f9f5a & 15 & 45 & 0.01194 & 2.854 & 239.1$\times$ \\
e4ecb462-e2be-41b3-8efc-795f49ef07bf & 15 & 82 & 0.01724 & 2.808 & 162.9$\times$ \\
6bdb38e6-2ce5-4f77-bef8-b87a7f0cc525 & 15 & 83 & 0.01740 & 4.165 & 239.4$\times$ \\
fb1ceff0-a905-45e7-8639-1c610d1bfbab & 15 & 84 & 0.01718 & 2.715 & 158.0$\times$ \\
786b5173-e297-4ec3-82fa-450e9c79fb7d & 15 & 89 & 0.01752 & 4.298 & 245.3$\times$ \\
cdc0ff86-2eaa-4d2a-905d-2e14ee189011 & 15 & 90 & 0.01703 & 4.346 & 255.1$\times$ \\
3eab2c37-dbd9-4f7a-890e-0643cc9ca2ee & 15 & 91 & 0.01838 & 2.886 & 157.0$\times$ \\
ed3e595b-e3ba-4049-9ee9-a51b460ceff5 & 16 & 33 & 0.01234 & 4.830 & 391.4$\times$ \\
9810dadf-1fe0-48ab-a1c1-d1e0e01f233e & 16 & 34 & 0.01205 & 3.085 & 256.0$\times$ \\
19e7663d-29d7-4f96-9262-9c7bc068fadd & 16 & 38 & 0.01245 & 4.030 & 323.6$\times$ \\
696dbfa4-e56c-4a09-a907-2abbad6de073 & 16 & 43 & 0.01298 & 3.024 & 233.0$\times$ \\
dba1e960-49e4-4152-8dce-f6da100c9561 & 25 & 30 & 0.002118 & 8.076 & 3813.4$\times$ \\
6e4e9b37-0d56-419b-ac54-69049066f914 & 26 & 30 & 0.002144 & 6.639 & 3096.4$\times$ \\
4a616af2-bd14-42f1-beed-a3c0ca68291e & 27 & 32 & 0.002357 & 4.479 & 1900.3$\times$ \\
34195ade-a79a-4c96-adeb-7221109fa2a4 & 27 & 82 & 0.01757 & 4.426 & 252.0$\times$ \\
8635db8f-5b62-464e-9b5b-0863d65788ee & 27 & 91 & 0.01771 & 4.378 & 247.2$\times$ \\
abc9d12c-2863-48db-b97f-35af7267ccf5 & 29 & 1 & 0.001856 & 4.536 & 2444.0$\times$ \\
37098ea3-9ce7-40bd-830c-b24999ce2434 & 29 & 18 & 0.002048 & 7.713 & 3766.0$\times$ \\
1571c14a-181c-4f15-97a5-178e4b316ca5 & 29 & 19 & 0.002080 & 4.573 & 2198.5$\times$ \\
03910df4-7c56-4612-8d08-0d82a681d0db & 29 & 30 & 0.002091 & 4.790 & 2291.0$\times$ \\
e667d2ac-c644-4abe-ba3b-7ab09b5b2dfd & 29 & 32 & 0.002331 & 7.635 & 3275.8$\times$ \\
e63194e7-b79e-4e50-b08a-16fd1f700eb5 & 29 & 36 & 0.01302 & 4.592 & 352.6$\times$ \\
fc14d852-74ad-41e7-a538-61b89fae5d9c & 29 & 44 & 0.01205 & 7.300 & 605.9$\times$ \\
27c3374f-c621-43f2-8a1a-8710f6ef22ba & 29 & 45 & 0.01254 & 4.744 & 378.2$\times$ \\
0c4f5578-a698-497e-95eb-873e2faf7df4 & 29 & 83 & 0.01741 & 4.835 & 277.7$\times$ \\
27afdcea-6ed1-434b-bf24-a7566c4b1a56 & 29 & 84 & 0.01740 & 5.059 & 290.7$\times$ \\
a876010b-d67f-492d-a7e6-76e9d59e5403 & 29 & 89 & 0.01833 & 7.864 & 429.2$\times$ \\
6b10b6da-7a58-4abc-9c40-d1fde49035d8 & 29 & 91 & 0.01759 & 4.574 & 260.0$\times$ \\
a30b4f8d-d6d7-4a9a-8a80-f85e3ec629a4 & 30 & 18 & 0.002074 & 7.591 & 3659.5$\times$ \\
e0488cb7-b07a-4019-ab24-4b2f97af2530 & 30 & 19 & 0.002091 & 5.043 & 2412.2$\times$ \\
6832006b-2f6b-42d1-9eeb-7ef92fec7d1a & 30 & 28 & 0.002075 & 4.861 & 2342.7$\times$ \\
4a0e0529-5a2b-4f3a-9675-3f87e10c5ad9 & 30 & 29 & 0.002112 & 6.481 & 3068.4$\times$ \\
7f20565a-5f0d-4625-b57f-4d3f6a52f0db & 30 & 30 & 0.002133 & 4.798 & 2249.1$\times$ \\
9f252ffa-e3d6-4cdb-b94e-a9840d272c57 & 30 & 31 & 0.002139 & 6.977 & 3262.2$\times$ \\
8ba75447-ae9e-4769-a5de-76e8d1685d6e & 30 & 32 & 0.002149 & 7.704 & 3584.4$\times$ \\
30a90fa5-375f-4bdb-aa48-79e59b5f0cac & 30 & 33 & 0.01291 & 4.866 & 377.0$\times$ \\
a03d722b-1858-43a9-86a0-37ea08cd7931 & 30 & 34 & 0.01204 & 5.085 & 422.5$\times$ \\
de54c4e6-7c89-43c7-aefb-db20265f4cdf & 30 & 35 & 0.01223 & 4.871 & 398.1$\times$ \\
2f3b7321-e55c-4a11-9ab3-4aab5dd4ab3a & 30 & 36 & 0.01260 & 4.725 & 374.9$\times$ \\
8bdd4f88-c992-4c75-b78b-b4dc626ca409 & 30 & 43 & 0.01245 & 4.886 & 392.3$\times$ \\
ee6946e7-b658-4e46-a389-7613624b3e78 & 30 & 44 & 0.01265 & 4.864 & 384.7$\times$ \\
6b4b9d2b-1cda-4ea7-8b4c-deca25596e70 & 30 & 82 & 0.01628 & 5.031 & 309.1$\times$ \\
f362edf4-723c-4751-afd9-5a0a6abbf095 & 30 & 83 & 0.01702 & 4.956 & 291.1$\times$ \\
bb0f8277-4d31-443b-9454-ac1a547750d3 & 30 & 89 & 0.01747 & 4.690 & 268.5$\times$ \\
22207643-eed4-49cf-a2a4-70877903ec3e & 30 & 90 & 0.01824 & 4.803 & 263.3$\times$ \\
5db1b172-eda8-4714-9981-a069dc33d7e9 & 30 & 91 & 0.01755 & 8.491 & 483.9$\times$ \\
bda73497-5c27-4f0a-b6b1-52ca3c2095d1 & 31 & 26 & 0.002075 & 4.932 & 2376.9$\times$ \\
a52c09bc-2ee5-4366-be02-457932a80631 & 31 & 43 & 0.01333 & 6.814 & 511.2$\times$ \\
\bottomrule
\end{longtable}
}

\Needspace{12\baselineskip}

\subsection{DSA Sparse Attention}

{\scriptsize
\setlength{\tabcolsep}{2pt}
\setlength{\abovecaptionskip}{2pt}
\setlength{\belowcaptionskip}{3pt}
\setlength{\LTpre}{2pt}
\setlength{\LTpost}{3pt}
\renewcommand{\arraystretch}{0.88}
\begin{longtable}{@{}>{\raggedright\arraybackslash}p{0.42\textwidth}>{\raggedleft\arraybackslash}p{0.13\textwidth}>{\raggedleft\arraybackslash}p{0.12\textwidth}>{\raggedleft\arraybackslash}p{0.12\textwidth}>{\raggedleft\arraybackslash}p{0.12\textwidth}@{}}
\caption{Per-workload retained results for DSA Sparse Attention. Latency and PyTorch reference are milliseconds; speedup is relative to PyTorch. Fixed: page count = 8462.}
\label{tab:appendix-shapes-dsa-sparse-attention}\\
\toprule
\textbf{Workload UUID} & \textbf{\shortstack{Token\\count}} & \textbf{Latency} & \textbf{\shortstack{PyTorch\\reference}} & \textbf{\shortstack{Speedup vs.\\PyTorch}} \\
\midrule
\endfirsthead
\multicolumn{5}{@{}l}{\textit{Table~\thetable{} continued from previous page}} \\
\toprule
\textbf{Workload UUID} & \textbf{\shortstack{Token\\count}} & \textbf{Latency} & \textbf{\shortstack{PyTorch\\reference}} & \textbf{\shortstack{Speedup vs.\\PyTorch}} \\
\midrule
\endhead
0c23b10c7b7645719517828c12eaa1d2 & 1 & 0.005248 & 1.044 & 199.0$\times$ \\
05f6de657db543ae9e4c46796522843a & 2 & 0.009899 & 1.621 & 163.8$\times$ \\
0a63b87bb2e54e9db1ca3c4c53a1d521 & 2 & 0.006763 & 1.243 & 183.7$\times$ \\
9d4a5f21268e484ea05a2f2af91d9fa7 & 2 & 0.005419 & 1.249 & 230.5$\times$ \\
9f3f891bfbe24776adfcd5a579d093dd & 2 & 0.009952 & 1.582 & 159.0$\times$ \\
b7668cfd194c4b95ab600feb205ebac6 & 2 & 0.006581 & 1.581 & 240.2$\times$ \\
e6b849f2900446148c01d81152efae23 & 2 & 0.007792 & 1.629 & 209.0$\times$ \\
f77df5ce21634f7ca0586af7e3c13434 & 2 & 0.005498 & 1.580 & 287.3$\times$ \\
fc85411e250c41879b6d6a1edb80f0a7 & 2 & 0.005419 & 1.263 & 233.1$\times$ \\
68d6817dcfd1433aa8d2ddeefd54b6ea & 6 & 0.007957 & 3.042 & 382.3$\times$ \\
d57eb9e19f0642e8af9bd76ad0823303 & 6 & 0.01527 & 3.149 & 206.2$\times$ \\
ddfa9e340b264f76abe7418692faa876 & 6 & 0.01500 & 2.097 & 139.8$\times$ \\
2207f0fdc96347c59d106e4976cfad57 & 7 & 0.01380 & 4.795 & 347.4$\times$ \\
3838996164a94d728710f913477feba8 & 7 & 0.01151 & 4.779 & 415.2$\times$ \\
ae4219a95f044f45bd10e17ab63c6e8f & 7 & 0.01356 & 3.259 & 240.3$\times$ \\
02d6ae9c64ab42ff93f05c23c53bcb7d & 8 & 0.01487 & 3.993 & 268.5$\times$ \\
232ed014bafc4835b9881bb308c659b0 & 8 & 0.01455 & 2.528 & 173.7$\times$ \\
385742b2717e4f02b918c7349dde23d8 & 8 & 0.01329 & 2.534 & 190.7$\times$ \\
4c46a94ba2364dc7ab476286dee8dce3 & 8 & 0.01302 & 2.619 & 201.2$\times$ \\
5096e459ce3f4cdf82773ce1a0c73c8a & 8 & 0.01511 & 2.525 & 167.0$\times$ \\
564007ac354e4662a62cc4d6352dc494 & 8 & 0.01530 & 2.598 & 169.8$\times$ \\
78b2e11c30004cceb84355722a7c6b0a & 8 & 0.01516 & 2.522 & 166.3$\times$ \\
7a389715dcc3479e9b0512309f9d1d56 & 8 & 0.01495 & 2.584 & 172.8$\times$ \\
\bottomrule
\end{longtable}
}

\Needspace{12\baselineskip}

\subsection{GDN Decode}

{\scriptsize
\setlength{\tabcolsep}{2pt}
\setlength{\abovecaptionskip}{2pt}
\setlength{\belowcaptionskip}{3pt}
\setlength{\LTpre}{2pt}
\setlength{\LTpost}{3pt}
\renewcommand{\arraystretch}{0.88}
\begin{longtable}{@{}>{\raggedright\arraybackslash}p{0.42\textwidth}>{\raggedleft\arraybackslash}p{0.13\textwidth}>{\raggedleft\arraybackslash}p{0.12\textwidth}>{\raggedleft\arraybackslash}p{0.12\textwidth}>{\raggedleft\arraybackslash}p{0.12\textwidth}@{}}
\caption{Per-workload retained results for GDN Decode. Latency and PyTorch reference are milliseconds; speedup is relative to PyTorch.}
\label{tab:appendix-shapes-gdn-decode}\\
\toprule
\textbf{Workload UUID} & \textbf{\shortstack{Batch\\size}} & \textbf{Latency} & \textbf{\shortstack{PyTorch\\reference}} & \textbf{\shortstack{Speedup vs.\\PyTorch}} \\
\midrule
\endfirsthead
\multicolumn{5}{@{}l}{\textit{Table~\thetable{} continued from previous page}} \\
\toprule
\textbf{Workload UUID} & \textbf{\shortstack{Batch\\size}} & \textbf{Latency} & \textbf{\shortstack{PyTorch\\reference}} & \textbf{\shortstack{Speedup vs.\\PyTorch}} \\
\midrule
\endhead
22d5cef5-4f30-4f43-9d5f-0e9e95dc2201 & 1 & 0.002470 & 1.030 & 416.2$\times$ \\
3daa0974-293c-4414-b3c2-1f04368c1189 & 1 & 0.002500 & 1.719 & 688.8$\times$ \\
49a125e5-edf0-492e-a8b3-3676d14adaa3 & 1 & 0.002470 & 1.166 & 471.0$\times$ \\
4c7df22f-70ef-4494-864f-6f10209ab0f3 & 1 & 0.002470 & 1.010 & 409.0$\times$ \\
5716e24a-3f55-411a-bcd8-e6b677b1ca7e & 1 & 0.002470 & 1.492 & 603.0$\times$ \\
798635cf-d424-4343-a959-c96b0c0e81fb & 1 & 0.002450 & 1.035 & 421.9$\times$ \\
901e5104-dccb-4c3f-ae13-ef4d31a4d456 & 1 & 0.002420 & 1.026 & 423.8$\times$ \\
a5714b69-525c-4b95-bb7a-a0f9770c2f48 & 1 & 0.002460 & 1.031 & 419.2$\times$ \\
aed4bdd4-3139-4a1b-ae2f-aab8d4ba4090 & 1 & 0.002440 & 1.022 & 419.4$\times$ \\
d0e91dea-aa1b-46c8-a67c-b2814f5a1725 & 1 & 0.002460 & 1.026 & 416.6$\times$ \\
26f760ab-c286-41bf-8c37-3bc5df4c98fc & 4 & 0.003020 & 3.572 & 1183.2$\times$ \\
494844dc-80e9-41e9-9fe2-3d2618fdef64 & 4 & 0.003060 & 3.596 & 1174.6$\times$ \\
6d929bff-d051-4e1d-acbf-7d0cc13f6dc8 & 4 & 0.003030 & 5.933 & 1958.2$\times$ \\
9a92acbc-6104-48c6-8176-601feee30001 & 4 & 0.003030 & 3.785 & 1249.4$\times$ \\
9f238670-9a56-4ab9-94f9-555755f32205 & 4 & 0.003030 & 3.692 & 1216.7$\times$ \\
deecf42e-dfae-4061-936f-0af4d892c231 & 4 & 0.003060 & 3.787 & 1237.1$\times$ \\
ec9d2340-6d13-40e4-a6fe-4483a1cacd0d & 4 & 0.003030 & 3.636 & 1200.3$\times$ \\
fa90b213-3ad2-4a42-9173-97fbf9e2e809 & 4 & 0.003060 & 3.527 & 1152.1$\times$ \\
25d1f606-3e9e-49e6-a3f0-991092c7a845 & 8 & 0.004030 & 7.267 & 1802.5$\times$ \\
2640f1e3-4b02-4041-bdc1-59a28e0b9954 & 8 & 0.004030 & 7.094 & 1759.4$\times$ \\
61dfd334-971b-4a44-a8a8-f00950192745 & 8 & 0.004020 & 6.988 & 1737.7$\times$ \\
7464a1d6-b0dd-4061-b15a-ebdabf47e351 & 8 & 0.004050 & 7.299 & 1800.8$\times$ \\
abec5d32-2409-4412-8683-472f3a091a54 & 8 & 0.004070 & 12.363 & 3034.1$\times$ \\
ca62112d-68e8-4eb7-a318-26c48d256e10 & 8 & 0.004090 & 11.835 & 2896.9$\times$ \\
fe2e6584-17d0-4025-ad44-db9e507f6bed & 8 & 0.004060 & 7.697 & 1893.9$\times$ \\
63d4cb66-d6f6-4ceb-8083-0c0fb4abece9 & 16 & 0.004910 & 24.079 & 4907.5$\times$ \\
76eec66d-0ada-4c35-bf21-a07247ad7f05 & 16 & 0.004950 & 14.144 & 2857.7$\times$ \\
ac83f1e4-982e-41a5-8c02-7ccc14c728d4 & 16 & 0.004930 & 13.973 & 2835.5$\times$ \\
d0acf04a-5919-4bf3-95ca-94fbef5786c0 & 16 & 0.004910 & 14.448 & 2944.5$\times$ \\
d1508f6d-81b7-4caf-947b-b48f612a3061 & 16 & 0.004910 & 13.750 & 2802.5$\times$ \\
ef6bb01d-1294-4a06-97b2-604ec944b4c0 & 16 & 0.004910 & 14.380 & 2930.6$\times$ \\
f0507025-1db0-48de-b3d6-4faad5632558 & 16 & 0.004910 & 24.224 & 4936.9$\times$ \\
0915952b-5887-43aa-9807-a0980e7a78f6 & 32 & 0.007430 & 42.777 & 5753.9$\times$ \\
388f79b1-13e9-44e6-b12e-c8197989a924 & 32 & 0.007460 & 76.714 & 10281.6$\times$ \\
40940a4c-1348-416d-aea1-f3183548229b & 32 & 0.007400 & 69.995 & 9462.3$\times$ \\
773a4157-01fa-4637-90be-8604e0b40526 & 32 & 0.007480 & 139.242 & 18621.9$\times$ \\
80288532-6967-4b8e-b5bc-c31ce0a208d5 & 32 & 0.007490 & 89.847 & 11990.2$\times$ \\
a8c8beff-e414-4580-b2f7-e5b8f13bc269 & 32 & 0.007410 & 37.646 & 5078.1$\times$ \\
d38dde79-79f9-4c2d-8bbe-05ee1be48583 & 32 & 0.007440 & 46.183 & 6207.4$\times$ \\
34403e5b-8551-4d44-95b0-21532c4eb839 & 48 & 0.01006 & 93.661 & 9306.6$\times$ \\
3803367a-f140-432a-9784-43fe5d5f64d0 & 48 & 0.01010 & 143.640 & 14219.7$\times$ \\
45e7697a-bae5-42bd-9f56-959c10ff681f & 48 & 0.01009 & 90.708 & 8989.3$\times$ \\
4c38f0e4-aabf-4742-90b9-24231effd96f & 48 & 0.01012 & 90.277 & 8923.3$\times$ \\
53385c7f-393d-41db-aec8-5b9eb5bf35d1 & 48 & 0.01012 & 146.110 & 14433.5$\times$ \\
b91cc4aa-048d-4e05-9c0c-8a153ec997f8 & 48 & 0.01010 & 87.311 & 8643.5$\times$ \\
be91d78c-1114-4563-b0aa-e0682ccc09bb & 48 & 0.01006 & 94.511 & 9396.0$\times$ \\
42963acb-f2f5-4ada-9205-3931cd26fa44 & 64 & 0.01257 & 128.819 & 10251.8$\times$ \\
58571e49-29ee-4a53-bf1d-a7a2363e9db3 & 64 & 0.01259 & 195.477 & 15524.1$\times$ \\
8d0c5789-550a-4216-a471-202b2655a4e8 & 64 & 0.01246 & 203.409 & 16327.1$\times$ \\
9eef9bc7-5972-437e-b488-853c6dacb470 & 64 & 0.01246 & 136.420 & 10949.6$\times$ \\
b1931bc6-b8b6-44da-a83b-aa6cc4d4c275 & 64 & 0.01276 & 139.802 & 10954.0$\times$ \\
eaf0a285-447c-4432-8e68-d287acc3cb08 & 64 & 0.01258 & 133.507 & 10616.0$\times$ \\
ef2cf980-6977-49d1-a6f8-7247becf0273 & 64 & 0.01251 & 144.541 & 11552.2$\times$ \\
f4fd2171-c869-440c-b199-f403e3c6b788 & 64 & 0.01248 & 115.454 & 9251.0$\times$ \\
\bottomrule
\end{longtable}
}

\Needspace{12\baselineskip}

\subsection{GDN Prefill}

{\scriptsize
\setlength{\tabcolsep}{2pt}
\setlength{\abovecaptionskip}{2pt}
\setlength{\belowcaptionskip}{3pt}
\setlength{\LTpre}{2pt}
\setlength{\LTpost}{3pt}
\renewcommand{\arraystretch}{0.88}
\begin{longtable}{@{}>{\raggedright\arraybackslash}p{0.31\textwidth}>{\raggedleft\arraybackslash}p{0.095\textwidth}>{\raggedleft\arraybackslash}p{0.095\textwidth}>{\raggedleft\arraybackslash}p{0.095\textwidth}>{\raggedleft\arraybackslash}p{0.11\textwidth}>{\raggedleft\arraybackslash}p{0.11\textwidth}>{\raggedleft\arraybackslash}p{0.11\textwidth}@{}}
\caption{Per-workload retained results for GDN Prefill. Latency and PyTorch reference are milliseconds; speedup is relative to PyTorch.}
\label{tab:appendix-shapes-gdn-prefill}\\
\toprule
\textbf{Workload UUID} & \textbf{\shortstack{Total sequence\\length}} & \textbf{\shortstack{Sequence\\count}} & \textbf{\shortstack{Cumulative\\length\\entries}} & \textbf{Latency} & \textbf{\shortstack{PyTorch\\reference}} & \textbf{\shortstack{Speedup vs.\\PyTorch}} \\
\midrule
\endfirsthead
\multicolumn{7}{@{}l}{\textit{Table~\thetable{} continued from previous page}} \\
\toprule
\textbf{Workload UUID} & \textbf{\shortstack{Total sequence\\length}} & \textbf{\shortstack{Sequence\\count}} & \textbf{\shortstack{Cumulative\\length\\entries}} & \textbf{Latency} & \textbf{\shortstack{PyTorch\\reference}} & \textbf{\shortstack{Speedup vs.\\PyTorch}} \\
\midrule
\endhead
77daf91d-0660-4c4b-8c32-336a69281cd9 & 6 & 1 & 2 & 0.005470 & 1.560 & 285.2$\times$ \\
8e7ef744-5f4e-40e7-aae3-471e013657bd & 12 & 1 & 2 & 0.008590 & 1.700 & 198.0$\times$ \\
685d26ff-9ad1-4378-b15a-b279d2db1bcb & 13 & 1 & 2 & 0.009090 & 1.906 & 209.7$\times$ \\
eaa0fd47-ac57-47d0-a896-e3b1a914a5da & 14 & 1 & 2 & 0.009560 & 3.971 & 415.5$\times$ \\
339a7ff4-0a55-40d0-99d1-f33e3b1ffe4a & 16 & 1 & 2 & 0.01069 & 2.627 & 245.7$\times$ \\
ef9515b6-ad88-4a3e-bd89-31384ddd53ad & 16 & 1 & 2 & 0.01094 & 2.201 & 201.1$\times$ \\
f622a11d-8b79-40b6-9949-5509afe827c8 & 16 & 1 & 2 & 0.01091 & 2.258 & 206.9$\times$ \\
85d7becb-2969-4cc8-9120-1ec26a78d258 & 18 & 1 & 2 & 0.01165 & 2.492 & 214.0$\times$ \\
cc241d2e-f8b7-448f-95ec-1d5d2809c068 & 23 & 1 & 2 & 0.01380 & 3.257 & 235.9$\times$ \\
28b70283-08fd-4eb9-8831-147abbf8d0f2 & 24 & 1 & 2 & 0.01579 & 3.397 & 215.2$\times$ \\
2ba465c0-7405-4bf1-ba4e-7d632d95f141 & 28 & 2 & 3 & 0.009970 & 4.264 & 427.6$\times$ \\
02c1e5f0-0d21-4c3c-a22d-40ad45a93549 & 30 & 1 & 2 & 0.01928 & 7.806 & 404.8$\times$ \\
62447caf-853a-4bf8-a62e-d12885248881 & 30 & 1 & 2 & 0.01754 & 4.092 & 233.3$\times$ \\
977d19f8-5d0f-4c19-91e0-b05b500b34fc & 30 & 1 & 2 & 0.01746 & 4.121 & 236.0$\times$ \\
d5f5c00c-c159-45c8-aa2a-2bb8a00e2fe8 & 31 & 1 & 2 & 0.01723 & 7.202 & 418.1$\times$ \\
043e74e4-0297-4988-808a-0090be48a90e & 32 & 2 & 3 & 0.01207 & 4.517 & 374.1$\times$ \\
d3dc3577-8a65-4937-a63a-4aa7204181a7 & 35 & 1 & 2 & 0.01802 & 4.736 & 262.9$\times$ \\
d8f4a9ae-d391-4f70-8069-f5deb510a2d1 & 35 & 1 & 2 & 0.01854 & 4.902 & 264.4$\times$ \\
1aa8cf18-0fca-4e32-a7ff-7ed29d24a3d2 & 35 & 2 & 3 & 0.01274 & 4.655 & 365.5$\times$ \\
f2f01c2c-3327-4cae-a0d6-a12ec0e94fc3 & 35 & 2 & 3 & 0.01283 & 4.881 & 380.4$\times$ \\
1cf8e175-9f87-4add-984a-6c87a380fe9d & 40 & 2 & 3 & 0.01434 & 5.666 & 395.2$\times$ \\
1efaf2a9-05db-4737-8f41-6880ba1bb487 & 42 & 2 & 3 & 0.01610 & 6.639 & 412.4$\times$ \\
fdf5f1f4-1135-4d3f-99e4-8965958317fc & 46 & 2 & 3 & 0.01692 & 6.262 & 370.1$\times$ \\
f203fdcd-6b72-4687-9fcf-a010066eab5c & 48 & 1 & 2 & 0.01693 & 11.495 & 679.1$\times$ \\
27f44fd6-0b0d-44bc-a771-617fd3e9e61c & 49 & 1 & 2 & 0.01870 & 6.490 & 347.1$\times$ \\
2c9693b4-6375-459a-a29f-650c24abe472 & 49 & 1 & 2 & 0.01890 & 6.764 & 357.8$\times$ \\
9343fd82-a06d-493a-9918-1044b0c1cbd1 & 49 & 1 & 2 & 0.01846 & 6.591 & 357.0$\times$ \\
a39aa135-dd03-4abb-862d-020cf52fc7dd & 61 & 1 & 2 & 0.01823 & 8.464 & 464.3$\times$ \\
49ef89d2-da00-4cd5-b912-708cbfc7106d & 67 & 1 & 2 & 0.02050 & 9.361 & 456.6$\times$ \\
056224b8-0a27-4634-b93b-36beb658e135 & 68 & 1 & 2 & 0.02062 & 9.305 & 451.3$\times$ \\
3a77dfec-f1c1-40aa-82e4-fafdfc84e540 & 76 & 2 & 3 & 0.01990 & 10.752 & 540.2$\times$ \\
cd979341-7112-4887-a2f3-dfb269a49bbc & 76 & 2 & 3 & 0.01835 & 10.406 & 567.2$\times$ \\
ea27be17-3dfa-423e-aa85-e9001c24daed & 76 & 2 & 3 & 0.01824 & 9.948 & 545.4$\times$ \\
7a7deca8-255f-4400-9fc6-367d4671ddcc & 79 & 1 & 2 & 0.02060 & 10.883 & 528.4$\times$ \\
078a41ea-3a59-4b4f-b3c2-95cb6e538a6e & 81 & 2 & 3 & 0.01930 & 10.459 & 542.1$\times$ \\
d2b5a221-f890-4cdd-83bc-bdbff1f0d0b0 & 82 & 3 & 4 & 0.01706 & 20.048 & 1175.4$\times$ \\
f3d30cb9-65d2-452a-92ef-10fb8b78473d & 85 & 2 & 3 & 0.02077 & 11.929 & 574.4$\times$ \\
19c6ab20-df55-4b2c-b9a7-d76870b50e3e & 91 & 2 & 3 & 0.02049 & 12.315 & 601.0$\times$ \\
1b441950-6f77-4565-bada-12294714fb26 & 91 & 2 & 3 & 0.02077 & 12.287 & 591.7$\times$ \\
a0eb2dc2-9de3-47fc-8ab4-0beee6150015 & 92 & 1 & 2 & 0.02094 & 16.876 & 806.0$\times$ \\
2683c087-e726-4a48-af25-674a95a889f2 & 97 & 1 & 2 & 0.02085 & 13.029 & 624.8$\times$ \\
73b8cc85-11ec-4ac4-8469-fce7b52ee44a & 97 & 1 & 2 & 0.02073 & 13.194 & 636.6$\times$ \\
35ea9bbe-0189-4089-878c-4fc21461297e & 114 & 2 & 3 & 0.02073 & 15.354 & 740.9$\times$ \\
fd072ba6-2190-4ce6-b96c-5212d4caf6a0 & 114 & 2 & 3 & 0.02071 & 17.085 & 824.8$\times$ \\
ed66c791-0393-45d5-946b-4cd5d3e8b395 & 116 & 1 & 2 & 0.02094 & 15.960 & 762.2$\times$ \\
33a38713-868b-43f7-8d75-fbe472e30387 & 118 & 2 & 3 & 0.02042 & 16.160 & 791.6$\times$ \\
25d9c14d-90ad-442d-8b0f-9452ad198832 & 132 & 2 & 3 & 0.02074 & 17.639 & 850.6$\times$ \\
ba08a83e-e151-4e16-bc70-abee6851604c & 134 & 1 & 2 & 0.02334 & 18.094 & 775.3$\times$ \\
c7846f96-c9ff-44d3-94ab-5eab99a1431b & 134 & 1 & 2 & 0.02306 & 18.091 & 784.5$\times$ \\
5d26ac5b-3c04-409d-a80d-5f1ac455916c & 139 & 3 & 4 & 0.02106 & 18.895 & 897.4$\times$ \\
fc7a2bcb-25d4-4840-a095-959499223267 & 139 & 3 & 4 & 0.02070 & 19.252 & 929.9$\times$ \\
ba95d412-06d5-47d9-99b8-185146c2f869 & 185 & 2 & 3 & 0.02345 & 24.585 & 1048.6$\times$ \\
87bff084-1b7a-478e-99fd-5952c989d80e & 202 & 3 & 4 & 0.02104 & 27.480 & 1305.8$\times$ \\
e92dafeb-3d50-4bce-90c3-6d0172e189d3 & 239 & 3 & 4 & 0.02080 & 34.179 & 1643.3$\times$ \\
bfd8f7b6-3953-46c3-af04-409b2fbe353c & 248 & 2 & 3 & 0.02556 & 34.531 & 1351.1$\times$ \\
d5aa60dc-f159-4d02-a1d4-f68bd2a62f33 & 255 & 3 & 4 & 0.02508 & 35.758 & 1425.9$\times$ \\
1d0cc342-9e3d-41e3-9ade-532551f86c17 & 294 & 3 & 4 & 0.02566 & 58.792 & 2290.8$\times$ \\
3215fe5f-4a3b-4eb6-af20-4e17368d87a9 & 341 & 2 & 3 & 0.02795 & 60.510 & 2165.2$\times$ \\
5a91aa02-8cee-49f9-aa2b-145d6e254ee4 & 401 & 4 & 5 & 0.02830 & 92.073 & 3253.7$\times$ \\
ce832e76-c4c2-421f-ad75-ebda2032c401 & 461 & 2 & 3 & 0.03273 & 289.221 & 8837.8$\times$ \\
352c9ace-2650-481f-ab51-69394fd0f366 & 525 & 1 & 2 & 0.03748 & 259.098 & 6912.5$\times$ \\
4b94d568-35ce-45a5-91eb-f8dc4b7077a7 & 574 & 2 & 3 & 0.03521 & 302.106 & 8579.9$\times$ \\
f105eda8-d047-4624-90a2-f5d5039e2f4e & 661 & 3 & 4 & 0.03329 & 318.857 & 9578.0$\times$ \\
6fbc155c-fcd0-4423-b1bf-970eef91d99b & 832 & 4 & 5 & 0.04268 & 349.354 & 8185.9$\times$ \\
43bf9699-adb3-4161-b741-8ff07b553336 & 959 & 4 & 5 & 0.05028 & 422.927 & 8411.2$\times$ \\
54856fec-3b50-4866-815d-19bfa16cf58b & 959 & 4 & 5 & 0.05005 & 408.706 & 8166.3$\times$ \\
08d4f2c4-dd07-4cc0-9d93-975c91b17577 & 973 & 3 & 4 & 0.05251 & 446.715 & 8506.9$\times$ \\
9c1ef562-578a-4df4-986e-e0c1aaa4727a & 973 & 3 & 4 & 0.05185 & 426.166 & 8219.3$\times$ \\
c5257f65-c411-4dbb-9dc1-ad4abcf00254 & 983 & 2 & 3 & 0.05251 & 457.248 & 8707.6$\times$ \\
f5619793-6267-4a96-8a23-6c6a937183e7 & 983 & 2 & 3 & 0.05250 & 488.885 & 9311.9$\times$ \\
f4926229-e4ba-49a3-8b2f-df2ec79c333b & 1377 & 1 & 2 & 0.06861 & 789.656 & 11509.7$\times$ \\
e44ba4d3-b2d7-4a2a-b860-0fcbdda6c351 & 1592 & 3 & 4 & 0.05975 & 816.945 & 13671.7$\times$ \\
e286a4f4-4a93-43bd-9465-b2915f0f8f96 & 1796 & 3 & 4 & 0.06184 & 1002.625 & 16212.0$\times$ \\
aaa378be-7365-4381-b2d3-35951ef88c7a & 1800 & 3 & 4 & 0.07370 & 932.093 & 12647.9$\times$ \\
c2bb4f66-6955-4e0f-b223-b802c1cd2348 & 1800 & 3 & 4 & 0.07381 & 941.103 & 12749.8$\times$ \\
c358edcd-3fda-43bb-98a2-ac32da9038b6 & 2040 & 2 & 3 & 0.09330 & 1111.892 & 11917.3$\times$ \\
6f1ad833-3d8a-4613-b5bd-a5d69263db1d & 2107 & 1 & 2 & 0.09610 & 1071.357 & 11148.9$\times$ \\
a01a3f93-19b0-4b12-95e3-6132a41d628e & 2284 & 3 & 4 & 0.09639 & 1520.513 & 15773.9$\times$ \\
a87ded8a-8202-405c-8ced-17521a8c8379 & 2857 & 3 & 4 & 0.07638 & 1636.886 & 21429.9$\times$ \\
15856e8c-d105-478d-b0de-4ef8141b5a77 & 3028 & 5 & 6 & 0.12028 & 1591.572 & 13232.6$\times$ \\
109addb1-15e0-4ff2-9b39-df3e79746af0 & 3271 & 2 & 3 & 0.11698 & 1790.491 & 15305.8$\times$ \\
7ba9d519-7c0b-4eac-b2ad-ae403dce2f95 & 3999 & 13 & 14 & 0.12561 & 2240.116 & 17833.8$\times$ \\
4b6143dd-0e5f-499f-93cb-076d9635bcd0 & 4124 & 15 & 16 & 0.06715 & 3820.843 & 56902.9$\times$ \\
07aa7922-1848-48a9-830a-54216b5553b3 & 5709 & 2 & 3 & 0.23132 & 3329.703 & 14394.6$\times$ \\
a9540651-862b-44af-ad75-57e2d9caf37a & 8192 & 20 & 21 & 0.15883 & 5356.494 & 33725.6$\times$ \\
cc310f94-60ea-4dcd-9e63-135dea91b83a & 8192 & 25 & 26 & 0.14417 & 5045.536 & 34997.4$\times$ \\
5835a2bc-8d60-43fc-b1ed-d4729ea62693 & 8192 & 32 & 33 & 0.16205 & 5111.205 & 31541.6$\times$ \\
d0ce7b5d-49e2-4a0b-b2ce-a087139b7d6b & 8192 & 32 & 33 & 0.12192 & 5038.906 & 41329.8$\times$ \\
5b8a0e4b-5ed7-45d0-aec3-c2c154e515b1 & 8192 & 34 & 35 & 0.16356 & 4763.918 & 29126.1$\times$ \\
5d3fc66a-6bba-4602-8319-a1da07045354 & 8192 & 34 & 35 & 0.16335 & 5067.068 & 31020.0$\times$ \\
618df04a-540c-42fd-b2ce-285dfdf78422 & 8192 & 35 & 36 & 0.12965 & 5056.808 & 39002.8$\times$ \\
c2931c92-bbe9-43d9-bc55-9aed77901266 & 8192 & 37 & 38 & 0.14784 & 5069.393 & 34290.0$\times$ \\
b8c8dc3c-3dfe-4334-a784-c477213e2f1a & 8192 & 38 & 39 & 0.12475 & 5306.986 & 42542.2$\times$ \\
e9e1e445-b41a-404b-a20f-bc869acc6cf0 & 8192 & 38 & 39 & 0.12450 & 5241.234 & 42098.2$\times$ \\
410794d4-6f70-4bb4-aed1-4669be9a610f & 8192 & 39 & 40 & 0.13467 & 5256.158 & 39031.2$\times$ \\
26244fb4-d295-4d4b-bf61-235d17fd0149 & 8192 & 43 & 44 & 0.15769 & 5120.555 & 32471.3$\times$ \\
a2629e02-611f-4264-8ee7-5c431f5fee4f & 8192 & 43 & 44 & 0.15743 & 5043.538 & 32037.0$\times$ \\
d49df0b2-7838-4865-9a51-7ec7877fe27f & 8192 & 48 & 49 & 0.13825 & 5094.162 & 36847.5$\times$ \\
06f21bb1-6cbd-4d55-b620-fb4d62181a71 & 8192 & 56 & 57 & 0.11425 & 5139.701 & 44986.0$\times$ \\
9a5d694b-7d4c-4ee6-8315-a13053ab6f92 & 8192 & 57 & 58 & 0.13454 & 5281.511 & 39256.7$\times$ \\
\bottomrule
\end{longtable}
}

\end{document}